\documentclass[preprint,12pt]{elsarticle}

\usepackage{acronym}%
\usepackage{amssymb}
\usepackage{tabularx}
\usepackage{array}
\usepackage{amsmath}
\usepackage{booktabs}
\usepackage{graphicx}
\usepackage{tikz}
\usepackage{adjustbox}
\usetikzlibrary{arrows.meta, positioning, fit, calc, backgrounds}

\usepackage{threeparttable}
\usepackage{makecell}
\newcolumntype{Y}{>{\raggedright\arraybackslash}X}
\newcommand{\best}[1]{\textbf{#1}}

\acrodef{ac:DL}[DL]{\emph{Deep learning}}
\acrodef{ac:ML}[ML]{\emph{Machine learning}}
\acrodef{ac:SSL}[SSL]{\emph{Self-supervised learning}}
\acrodef{ac:EF}[EF]{\emph{ejection fraction}}
\acrodef{ac:LVEF}[LVEF]{\emph{left-ventricular ejection fraction}}
\acrodef{ac:A2C}[A2C]{\emph{apical two-chamber}}
\acrodef{ac:A4C}[A4C]{\emph{apical four-chamber}}
\acrodef{ac:LV}[LV]{\emph{left ventricle}}
\acrodef{ac:ED}[ED]{\emph{end-diastole}}
\acrodef{ac:ES}[ES]{\emph{end-systole}}
\acrodef{ac:BYOL}[BYOL]{\emph{Bootstrap Your Own Latent}}
\acrodef{ac:BYOS}[BYOS]{\emph{Bootstrap Your Own Segmentation}}
\acrodef{ac:DINO}[DINO]{\emph{self-DIstillation with NO labels}}
\acrodef{ac:RAPTOR}[RAPTOR]{\emph{Robust And Parcelised Training Of echo Recordings}}
\acrodef{ac:STRAPTOR}[STRAPTOR]{\emph{Sequence-based Transformer for \ac{ac:RAPTOR}}}
\acrodef{ac:ViT}[ViT]{\emph{Vision Transformer}}
\acrodef{ac:IoU}[IoU]{\emph{intersection-over-union}}
\acrodef{ac:MAE}[MAE]{\emph{Mean Absolute Error}}
\acrodef{ac:BCE}[BCE]{\emph{Binary Cross-Entropy}}
\acrodef{ac:FT}[FT]{\emph{fine-tuning}}
\acrodef{ac:MLP}[MLP]{\emph{MultiLayer Perceptron}}

\journal{Medical Image Analysis}

\begin{document}

\begin{frontmatter}

\title{Evaluating self-supervised echocardiographic representations across downstream extraction strategies for left-ventricular segmentation and ejection fraction estimation}
\author[1]{Sylwia Majchrowska}

\affiliation[1]{organization={R\&D Data Science Skills \& Partnership, Data Science \& AI, BioPharma R\&D, AstraZeneca},
    addressline={Pepparedsleden 1}, 
    city={Mölndal},
    postcode={431 83}, 
    country={Sweden}
    }

\author[2]{Philip Teare}

\ead{philip.teare@astrazeneca.com}

\affiliation[2]{organization={Centre for AI, Data Science \& AI, BioPharma R\&D, AstraZeneca},
    addressline={Biomedical Campus, 1 Francis Crick Ave, Trumpington}, 
    city={Cambridge},
    postcode={CB2 0AA}, 
    country={UK}
    }

\begin{abstract}
Self-supervised learning (SSL) is increasingly used in medical imaging to reduce annotation requirements, but representation quality is often judged using a single downstream evaluation setting. For dense clinical tasks, this can confound representation quality with the capacity of the downstream model used to recover task-relevant information. We present a systematic evaluation of self-supervised representations for left-ventricular segmentation and ejection fraction (EF) estimation from apical four-chamber echocardiography on EchoNet-Dynamic. Rather than relying on a single downstream probe, we compare a hierarchy of extraction strategies with increasing expressivity: heuristic extraction without mask-supervised training, frozen linear probes, frozen lightweight decoder probes, and partial fine-tuning. We apply this framework to two complementary representation families: generic frozen self-DIstillation with NO labels (DINOv3) features and a task-adapted dense self-supervised representation, Bootstrap Your Own Segmentation (BYOS). In both families, heuristic extraction substantially understated what was recoverable from the frozen representation. For DINOv3, performance improved from Dice 0.684 and EF mean absolute error (MAE) 13.01 under heuristic extraction to Dice 0.906 and EF MAE 9.65 with a frozen lightweight decoder, approaching a supervised U-Net baseline (Dice 0.915, EF MAE 9.72). For BYOS, performance improved from Dice 0.687 and EF MAE 17.83 under heuristic extraction to Dice 0.902 and EF MAE 8.74 with a frozen lightweight decoder. Partial fine-tuning yielded only modest additional changes relative to strong frozen-decoder baselines. These results show that conclusions about self-supervised representation quality in dense echocardiographic analysis depend strongly on the downstream extraction strategy used for evaluation. We therefore argue that multi-strategy evaluation is an important methodological consideration for SSL in dense medical image analysis.
\end{abstract}

\begin{keyword}
Echocardiography \sep Ejection Fraction \sep Segmentation \sep Self-supervised learning

\end{keyword}

\end{frontmatter}

\section{Introduction}
\label{sec:introduction}

Quantitative echocardiography is central to routine cardiac imaging, and left-ventricular ejection fraction (\ac{ac:LVEF}) remains one of the most widely used measures of systolic function in both clinical practice and cardiovascular research \cite{lang2015recommendations}. Yet reliable estimation of \ac{ac:LVEF} from two-dimensional echocardiography remains difficult. Ultrasound images are affected by speckle, attenuation, acoustic shadowing, variable acquisition quality, and substantial operator dependence, while ventricular boundaries are often weak or ambiguous \cite{noble2006ultrasound, lang2015recommendations}. These factors complicate both chamber delineation and downstream functional measurement, and contribute to variability in automated and manual analysis alike.

Deep learning has improved performance on echocardiographic tasks such as chamber segmentation and direct \ac{ac:EF} estimation, particularly since the release of large public benchmarks such as EchoNet-Dynamic \cite{lit:EchoNetDynamic}. However, fully supervised approaches depend on expert annotations that are expensive to obtain at scale, especially for dense tasks such as frame-level segmentation. This makes echocardiography a natural setting for self-supervised learning (\ac{ac:SSL}), where large collections of unlabeled cine loops can be used to learn transferable visual representations before adaptation to downstream tasks \cite{azizi2021big, zhou2023self}.

Recent \ac{ac:SSL} methods, including contrastive, non-contrastive, and self-distillation approaches, have shown that strong representations can be learned without manual labels \cite{chen2020simclr, grill2020byol, caron2021dino, simeoni2025dinov3}. In medical imaging, these methods have improved label efficiency and downstream transfer across a range of modalities and tasks \cite{azizi2021big, zhou2023self}. However, for dense clinical prediction, an important methodological question remains unresolved: \emph{how should the quality of a self-supervised representation be evaluated?}

In much of the \ac{ac:SSL} literature, representation quality is inferred from performance under a single downstream adaptation setting, such as linear probing, one selected decoder architecture, or a heuristic extraction rule. For dense prediction tasks, this can be problematic. Downstream performance reflects not only the information present in the representation, but also the capacity of the downstream model used to recover that information. A weak result may therefore indicate a weak representation, an underpowered probe, or a mismatch between the representation and the extraction strategy used to evaluate it. As a consequence, evaluation with only one downstream strategy can give an incomplete or overly pessimistic view of representation quality.

This issue is especially relevant in echocardiography. Compared with many natural-image benchmarks, cardiac ultrasound presents low signal-to-noise ratio, weak and spatially ambiguous anatomical boundaries, substantial appearance variation across patients and acquisitions, and frequent visual continuity between adjacent chambers \cite{noble2006ultrasound, lit:EchoNetDynamic, ferreira2025label}. In such settings, a frozen representation may contain clinically useful anatomical structure without making that structure directly recoverable by a chamber-unspecific heuristic rule or a minimal linear probe. For dense echocardiographic analysis, the choice of downstream extraction strategy may therefore substantially influence the conclusions drawn about representation quality.

In this work, we investigate that question through a systematic evaluation of self-supervised representations for left-ventricular (\ac{ac:LV}) segmentation and \ac{ac:EF} estimation from apical four-chamber echocardiography. Rather than judging a representation under a single downstream model, we compare a controlled hierarchy of extraction strategies with increasing expressivity: heuristic extraction without mask-supervised training, frozen linear probes, frozen lightweight decoder probes, and partial \ac{ac:FT}. These are complemented by supervised segmentation and functional reference baselines. The purpose of this design is to distinguish limitations of the learned representation from limitations of the downstream mechanism used to interrogate it.

We apply this framework to two complementary representation families. The first is a generic frozen self-supervised foundation representation based on DINOv3 \cite{simeoni2025dinov3}. The second is a task-adapted dense self-supervised representation, termed \ac{ac:BYOS}, that combines a frozen DINOv3 backbone with an ultrasound-adapted trainable adapter, a U-Net-style decoder \cite{RonnebergerFB15}, and a \ac{ac:BYOL}-style dense consistency objective applied to a late spatial representation \cite{grill2020byol}. Studying both families allows us to test whether the main conclusions are specific to generic pretrained features or persist under a more task-adapted dense representation.

Using EchoNet-Dynamic \cite{lit:EchoNetDynamic}, we show that heuristic extraction substantially understates the utility of strong frozen self-supervised representations in this setting. For both DINOv3 and \ac{ac:BYOS}, performance improves markedly as the downstream extraction strategy becomes more expressive, \linebreak with lightweight decoder probes recovering substantially stronger \ac{ac:LV} structure than heuristic or linear approaches alone. Partial \ac{ac:FT} yields only modest additional changes relative to strong frozen-decoder baselines. We further show that segmentation overlap and functional accuracy are related but not equivalent, and that direct \ac{ac:EF} prediction remains stronger when explicit anatomical masks are not required.

The central contribution of this work is methodological: for dense echocardiographic prediction, conclusions about self-supervised representation quality depend strongly on the downstream extraction strategy used for evaluation. To address this, we introduce a structured evaluation framework spanning heuristic extraction, frozen linear probing, frozen lightweight decoding, partial \ac{ac:FT}, and supervised functional references. Using this framework, we show that weak heuristic or minimal probes can materially understate what is recoverable from a frozen representation, whereas lightweight decoder probes reveal substantially stronger anatomical and functional utility. By studying both generic DINOv3 features and a task-adapted dense self-supervised representation, we further show that this dependence on extraction strategy is not confined to a single representation design. Taken together, these findings support extraction-strategy-aware, multi-strategy evaluation as a more reliable basis for interpreting dense \ac{ac:SSL} representations in echocardiography.

\section{Related work}
\label{sec:related_work}

Deep learning has become an important tool for automated echocardiographic analysis, including chamber segmentation, view classification, and direct prediction of functional indices such as \ac{ac:EF}. A major catalyst for this line of work was the introduction of EchoNet-Dynamic, which provided a large public benchmark for \ac{ac:LV} segmentation and \ac{ac:EF} estimation from echocardiographic video \cite{lit:EchoNetDynamic}. Since then, supervised image- and video-based models have shown that strong performance is achievable for segmentation and functional prediction, while also highlighting persistent challenges related to sparse frame-level labels, image quality variability, view dependence, and temporal instability of framewise predictions. In segmentation-derived \ac{ac:EF} pipelines, an additional challenge is that strong framewise overlap does not necessarily imply accurate functional estimation: small contour errors, basal leakage, chamber confusion, and inconsistent temporal behavior can all distort area- or volume-derived \ac{ac:EF} estimates despite high Dice scores.

\ac{ac:SSL} has emerged as a powerful approach for learning transferable visual representations without manual annotation. In natural-image vision, contrastive, non-contrastive, and self-distillation methods such as SimCLR, \ac{ac:BYOL}, and \ac{ac:DINO} have shown that semantically rich representations can be learned directly from image data \cite{chen2020simclr, grill2020byol, caron2021dino}. More recently, DINOv3 demonstrated that large-scale self-supervised training can produce robust general-purpose visual features that transfer strongly across tasks and domains \cite{simeoni2025dinov3}. In medical imaging, \ac{ac:SSL} has increasingly been used to improve label efficiency and robustness across modalities, with reviews and benchmark studies reporting consistent downstream benefits, particularly when annotations are limited \cite{azizi2021big, zhou2023self}. However, most of this literature focuses primarily on whether self-supervised pretraining helps, rather than on how the learned representation should itself be evaluated in dense clinical settings.

Echocardiography is a particularly attractive application area for \ac{ac:SSL} because cine loops are abundant in routine care while dense expert annotations are relatively scarce. Recent work has explored self-supervised, weakly supervised, and label-efficient strategies for echocardiographic representation learning, including temporal pretext tasks, masked video modeling, distillation-based learning, and semi-supervised segmentation. These approaches exploit the structured spatiotemporal nature of cardiac motion while aiming to reduce reliance on manual contour annotation. Related work has also explored label-free or annotation-minimal cardiac ultrasound segmentation more directly. For example, Ferreira et al. proposed a self-supervised framework for label-free segmentation in cardiac ultrasound that combines weak spatial priors, iterative pseudo-label refinement, and deep learning to reduce dependence on manual annotation \cite{ferreira2025label}. In other work, authors combined DINO features with an unsupervised segmentation framework to generate interpretable echocardiographic chamber segmentations and downstream view classification without dense quantitative annotation \cite{majchrowska2025raptor}. Taken together, these studies highlight the practical value of unlabeled echocardiographic data, but also raise a methodological question that remains insufficiently resolved: when downstream extraction is weak or highly constrained, does poor performance reflect a weak representation, or simply an underpowered way of interrogating it?

This question connects directly to the broader literature on probing learned representations. In self-supervised vision, linear probing has become a standard tool for evaluating whether useful information is directly decodable from frozen features. While highly informative, linear probing provides only a partial view of representation quality, because it tests direct linear separability rather than the full amount of task-relevant structure that may be recoverable under modest downstream processing. Analyses of self-supervised vision transformers have shown that different forms of semantic and structural information may be distributed unevenly across layers and may not be equally visible under a single probe \cite{park2023what}. This matters especially for dense prediction tasks, where success depends not only on class separability but also on local spatial organization, multiscale structure, and boundary precision.

These concerns are particularly relevant for medical image segmentation. Dense \ac{ac:SSL} methods for medical imaging have emphasized the importance of spatially structured representations and dense downstream decoding, rather than relying only on global embeddings or weak image-level probes \cite{seince2024dense}. More broadly, frozen-feature transfer studies in dense prediction have shown that useful anatomical information can remain latent in pretrained representations even when weak downstream mappings understate it. For clinically structured tasks such as chamber segmentation, the representation may contain information that is not immediately accessible through a heuristic rule or a pointwise linear map, but becomes recoverable once limited nonlinear spatial decoding is introduced. This suggests that evaluation based on a single downstream probe can be actively misleading in dense settings, especially when the anatomy is weakly bounded or when neighboring structures have similar appearance.

Echocardiography is an especially demanding case for this problem. Compared with many natural-image benchmarks, cardiac ultrasound combines low signal-to-noise ratio, variable image quality, weak or discontinuous boundaries, and frequent visual continuity between adjacent chambers. Under these conditions, a frozen representation may encode useful anatomical structure without making it cleanly accessible through a chamber-unspecific heuristic rule or a minimal linear probe. As a result, conclusions about representation quality may depend strongly on the downstream strategy used to recover task-relevant information. This dependence is likely to be even more important when the target endpoint is not only anatomical overlap but also a derived functional quantity such as \ac{ac:EF}, for which small contour or temporal inconsistencies can propagate into clinically meaningful error.

A related line of work concerns how pretrained representations are adapted to downstream tasks. Full \ac{ac:FT} is often effective but can be computationally expensive and may obscure how much task-relevant information was already present in the frozen representation. For this reason, many recent studies compare frozen transfer with restricted adaptation or parameter-efficient tuning regimes, especially when large pretrained backbones are involved. In the present context, comparing heuristic extraction, frozen probes, and limited \ac{ac:FT} is useful not only as an engineering choice, but as an evaluation tool: it helps distinguish information that is already recoverable from the representation from information that only becomes available after supervised adaptation of the backbone.

Against this background, the present study makes two main contributions to the literature. First, it frames the evaluation of self-supervised echocardiographic representations itself as a methodological problem, rather than treating downstream probing as a fixed implementation detail. Second, it addresses this problem through a structured hierarchy of downstream extraction strategies spanning heuristic extraction, frozen supervised probes, partial \ac{ac:FT}, and functional reference baselines. This differs from most prior work, which typically evaluates a representation under only one selected probe or adaptation regime. In this sense, the present paper is not only about achieving strong \ac{ac:LV} segmentation and \ac{ac:EF} results on EchoNet-Dynamic, but also about clarifying how the quality of self-supervised representations should be interpreted for dense clinical prediction. Our central claim is that weak downstream extraction can substantially understate what is present in a learned representation, and that this issue should be treated as an integral part of evaluation design rather than as a secondary implementation choice.

\section{Methods}
\label{sec:methods}

\subsection{Study design and evaluation setting}
We investigated how self-supervised visual representations should be evaluated for \ac{ac:LV} segmentation and \ac{ac:EF} estimation in apical four-chamber echocardiography. The central aim was to distinguish limitations of the \emph{representation} from limitations of the downstream \emph{extraction strategy} used to recover task-relevant information. To do this, we compared a controlled hierarchy of downstream evaluation settings with increasing supervision and expressivity: heuristic extraction without mask-supervised training, frozen supervised probes, partial \ac{ac:FT}, and fully supervised reference baselines.

The purpose of this hierarchy was not to optimize the strongest possible method independently within each supervision regime, but to compare progressively more expressive ways of accessing the same underlying representation. In this sense, the evaluated settings should be interpreted as controlled downstream extraction regimes rather than as an exhaustive search over all possible task-specific models.

All segmentation-derived experiments were evaluated on the \emph{EchoNet-Dynamic} test split, while train and validation splits were used for model fitting and model selection. Because the dataset provides only sparse traced frames per video, segmentation accuracy was measured on labeled \ac{ac:ES} and \ac{ac:ED} frames using Dice and \ac{ac:IoU}. Functional performance was assessed by converting predicted masks into a video-level \ac{ac:EF} estimate and comparing it against the reference \ac{ac:EF} value using \ac{ac:MAE}. In addition, we compared \linebreak segmentation-derived methods with a direct \ac{ac:EF} regressor and with a ceiling obtained by applying the same \ac{ac:EF} proxy to ground-truth masks.

\subsection{Overview of method families}
The study compared five method families designed to separate the quality of the learned \emph{representation} from the capacity of the downstream \emph{extraction strategy}.

First, we evaluated \emph{heuristic extraction} methods, in which frozen self-supervised features were converted into masks by hand-designed rules without mask-supervised training. These baselines represent the most restricted downstream access regime and test whether \ac{ac:LV} structure can be recovered from the representation without learning a segmentation model. We considered both the \ac{ac:BYOS} hybrid representation and frozen \ac{ac:DINO}v3 features in this setting.

Second, we evaluated \emph{frozen supervised probes}, in which the representation was kept fixed and only a small supervised downstream model was trained using traced frames. We used both \emph{linear probes} and \emph{lightweight decoder probes}. Linear probes test whether \ac{ac:LV} structure is directly and approximately linearly separable in the frozen feature space. Lightweight decoder probes test whether stronger performance can be recovered by adding limited nonlinear spatial decoding while leaving the underlying representation unchanged.

Third, we evaluated \emph{partial \ac{ac:FT}}, in which a lightweight supervised decoder was trained jointly with a restricted subset of backbone or adapter parameters while the remainder of the model remained frozen. These experiments test whether modest task-specific adaptation provides a meaningful gain over the strongest frozen-feature baselines, or whether the frozen representation already contains most of the recoverable \ac{ac:LV} information.

Fourth, we included a \emph{fully supervised U-Net} baseline as the main mask-supervised reference. This baseline provides a practical upper reference for learned segmentation overlap under standard full supervision.

Fifth, we included two \emph{functional reference baselines}. The first was a \emph{direct EF regressor}, which predicts \ac{ac:EF} without producing explicit segmentation masks and therefore tests whether segmentation is necessary for best functional performance. The second was a \emph{ground-truth-mask EF ceiling}, obtained by applying the same \ac{ac:EF} proxy to oracle masks. This ceiling isolates how much of the remaining \ac{ac:EF} error arises from imperfect segmentation and frame selection rather than from the proxy itself.

\begin{figure*}[!h]
\centering
\resizebox{\textwidth}{!}{%
\begin{tikzpicture}[
    font=\small,
    >=Latex,
    method/.style={
        draw,
        rounded corners,
        align=center,
        minimum height=12mm,
        text width=28mm,
        inner sep=5pt
    },
    smallmethod/.style={
        draw,
        rounded corners,
        align=center,
        minimum height=12mm,
        text width=24mm,
        inner sep=4pt
    },
    paneltitle/.style={
        align=center,
        font=\bfseries\small,
        text width=38mm
    },
    panelsubtitle/.style={
        align=center,
        font=\bfseries\footnotesize,
        text width=26mm
    },
    paneltext/.style={
        align=center,
        font=\footnotesize,
        text width=38mm
    },
    widepaneltext/.style={
        align=center,
        font=\footnotesize,
        text width=58mm
    },
    panelbox/.style={
        draw,
        rounded corners,
        thick,
        inner sep=10pt
    },
    bluebox/.style={method, fill=blue!8},
    orangebox/.style={method, fill=orange!10},
    redbox/.style={method, fill=red!10},
    violetbox/.style={method, fill=violet!12},
    blueboxsmall/.style={smallmethod, fill=blue!8},
    orangeboxsmall/.style={smallmethod, fill=orange!10},
    arrow/.style={->, thick}
]

\node[paneltitle] (p1title) at (0,5.0) {Heuristic extraction};
\node[bluebox]   (m1) at (0,3.2) {BYOS heuristic};
\node[orangebox] (m2) at (0,1.2) {DINOv3 heuristic};
\node[paneltext] (p1text) at (0,-1.2) {Tests whether frozen SSL features can be converted into LV masks without supervised segmentation training.};
\node[panelbox, fit=(p1title)(m1)(m2)(p1text)] (P1) {};

\node[paneltitle] (p2title) at (7.8,5.0) {Frozen supervised probes};

\node[panelsubtitle] (p2a) at (6.2,3.9) {Linear probes};
\node[blueboxsmall]   (m4) at (6.2,2.5) {BYOS + linear probe};
\node[orangeboxsmall] (m5) at (6.2,0.7) {DINOv3 + linear probe};

\node[panelsubtitle] (p2b) at (9.4,3.9) {Lightweight decoders};
\node[blueboxsmall]   (m6) at (9.4,2.5) {BYOS + decoder};
\node[orangeboxsmall] (m7) at (9.4,0.7) {DINOv3 + decoder};

\node[widepaneltext] (p2text) at (7.8,-1.6) {Tests whether LV structure in frozen representations is directly linearly decodable or instead requires limited nonlinear spatial decoding for recovery.};

\node[panelbox, fit=(p2title)(p2a)(m4)(m5)(p2b)(m6)(m7)(p2text)] (P2) {};

\node[paneltitle] (p3title) at (15.2,5.0) {Partial fine-tuning\\(limited backbone adaptation)};
\node[bluebox]   (m8) at (15.2,3.1) {Partial FT BYOS + decoder};
\node[orangebox]   (m9) at (15.2,1.1) {Partial FT DINOv3 + decoder};
\node[paneltext] (p3text) at (15.2,-1.5) {Tests whether modest task-specific adaptation improves beyond strong frozen decoder probes.};
\node[panelbox, fit=(p3title)(m8)(m9)(p3text)] (P3) {};

\node[paneltitle] (p4title) at (21.6,5.0) {Direct EF prediction};
\node[redbox] (m10) at (21.6,2.6) {Direct EF regressor\\EF only};
\node[paneltext] (p4text) at (21.6,-0.5) {Best learned EF baseline when anatomical masks are not required.};
\node[panelbox, fit=(p4title)(m10)(p4text)] (P4) {};

\node[paneltitle] (p5title) at (27.0,5.0) {Ceiling};
\node[violetbox] (m11) at (27.0,2.6) {GT-mask EF proxy ceiling};
\node[paneltext] (p5text) at (27.0,-0.5) {Upper reference showing how much error remains when segmentation masks are perfect.};
\node[panelbox, fit=(p5title)(m11)(p5text)] (P5) {};

\draw[arrow] (P1.east) -- (P2.west);
\draw[arrow] (P2.east) -- (P3.west);
\draw[arrow] (P3.east) -- (P4.west);
\draw[arrow] (P4.east) -- (P5.west);

\end{tikzpicture}%
}
\caption{Overview of evaluated method families. The study compares heuristic extraction, frozen supervised probes, partial \ac{ac:FT}, direct EF regression, and a ground-truth-mask EF ceiling. The central methodological question is how the apparent value of a self-supervised representation changes as the downstream extraction strategy becomes more expressive.}
\label{fig:method_overview}
\end{figure*}
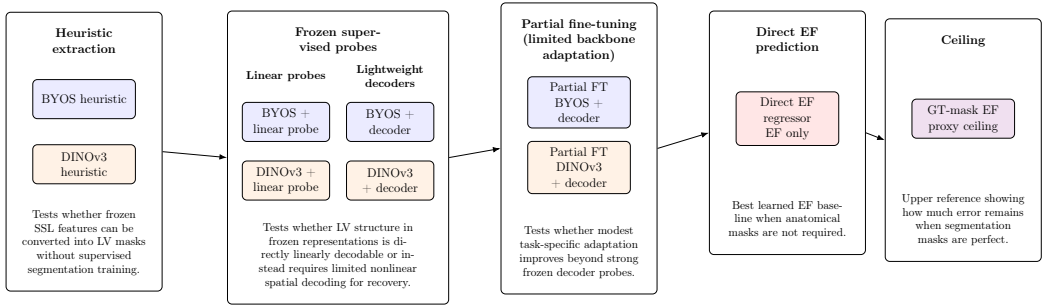

Together, these method families form a controlled hierarchy of downstream evaluation settings, from heuristic extraction without supervision to frozen probing, limited adaptation, and full supervision. This design allows us to distinguish limitations of the representation itself from limitations of the downstream mechanism used to recover task-relevant information. Figure~\ref{fig:method_overview} summarizes the evaluation setup. The figure should be read primarily as a hierarchy of increasing downstream expressivity, rather than as a ranking of methods.

\subsection{Hybrid self-supervised BYOS representation}
\label{subsec:byos}

Our principal self-supervised representation, referred to here as \ac{ac:BYOS}, was a hybrid dense self-supervised framework that combined a generic frozen foundation model with ultrasound-specific adaptation and dense self-distillation. 

The architecture consisted of a frozen DINOv3 ViT-S/16 backbone \cite{simeoni2025dinov3}, initialized with publicly released general-image weights, followed by a trainable adapter and a U-Net-style decoder. Concretely, the frozen DINOv3 ViT-S/16 backbone provided the final patch-token representation from \linebreak \texttt{get\_intermediate\_layers} with one extracted layer. For $512 \times 512$ inputs and patch size 16, these tokens were reshaped to a $32 \times 32$ spatial grid with embedding dimension 384. The trainable adapter first applied a $1 \times 1$ convolution to the token grid and then three successive $3 \times 3$ stride-2 convolutions, all with 384 channels, to construct a four-scale feature pyramid. The decoder was U-Net-like, with a center block followed by four upsampling stages with channel widths $(256,128,64,32)$. Each decoder block comprised two $3 \times 3$ convolution--batch-normalization--ReLU layers, with nearest-neighbor upsampling between stages and skip concatenation from the corresponding adapter feature map. The final decoder layer used a $1 \times 1$ convolution to produce the dense self-supervised representation. In the selected model, this dense representation had 5 output channels at full spatial resolution.

For the self-supervised objective, we attached a two-layer projection \ac{ac:MLP} and a two-layer predictor \ac{ac:MLP} to the flattened dense representation. Both used hidden dimension 2048, and the projection output dimension was 256. Batch normalization and ReLU nonlinearity were used in the hidden layers. The DINOv3 backbone remained frozen throughout BYOS pretraining, and only the adapter, decoder, projector, and predictor were optimized.
Unlike standard global self-supervised formulations, the objective was applied to a late dense decoder feature map rather than to a globally pooled embedding. This design was intended to encourage spatially coherent features that remain stable under augmentation and are more suitable for downstream dense prediction.

Given an input frame $x$, two stochastic augmented views $t_1(x)$ and $t_2(x)$ were generated and passed through the online encoder. A projection head and predictor were attached to the selected dense representation, and a momentum-updated target encoder provided the self-supervised target. The training objective followed the \ac{ac:BYOL} paradigm
\begin{equation}
\mathcal{L}_{\mathrm{BYOL}} =
\frac{1}{2}
\left[
\ell\!\left(q_{\theta}(z_{\theta}(t_1(x))), z_{\xi}(t_2(x))\right)
+
\ell\!\left(q_{\theta}(z_{\theta}(t_2(x))), z_{\xi}(t_1(x))\right)
\right],
\end{equation}
where $z_{\theta}$ denotes the online dense representation, $q_{\theta}$ the predictor, $z_{\xi}$ the target representation, and $\ell$ the cosine-similarity loss. In our implementation, the self-supervised objective was applied to the output of the decoder's final $1 \times 1$ convolution, prior to any downstream post-activation interpretation.

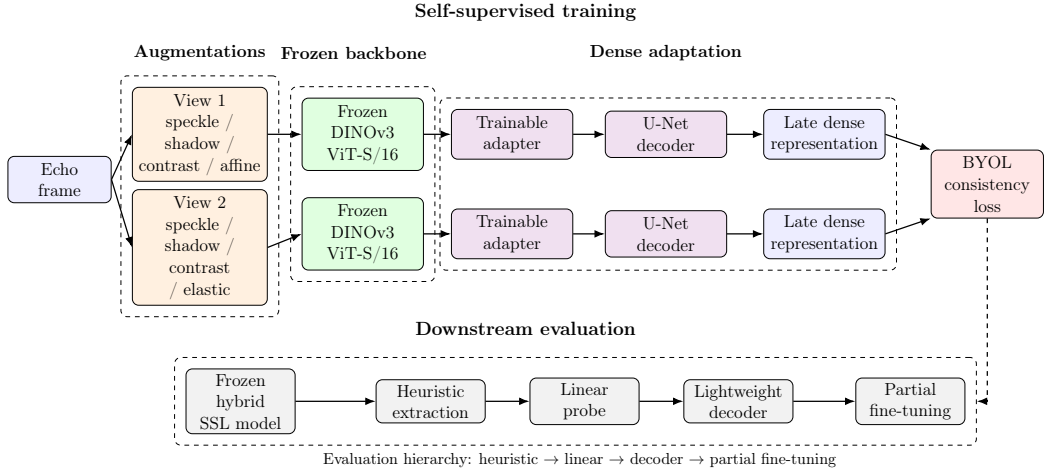
\begin{figure*}[!h]
\centering
\resizebox{\textwidth}{!}{%
\begin{tikzpicture}[
    font=\small,
    >=Latex,
    box/.style={
        draw,
        rounded corners,
        align=center,
        minimum height=9mm,
        text width=24mm,
        inner sep=4pt
    },
    boxp/.style={
        draw,
        rounded corners,
        align=center,
        minimum height=9mm,
        text width=20mm,
        inner sep=4pt
    },
    boxa/.style={
        draw,
        rounded corners,
        align=center,
        minimum height=9mm,
        text width=27mm,
        inner sep=4pt
    },
    smallbox/.style={
        draw,
        rounded corners,
        align=center,
        minimum height=8mm,
        text width=22mm,
        inner sep=3pt
    },
    proc/.style={boxp, fill=blue!7},
    procs/.style={box, fill=blue!7},
    aug/.style={boxa, fill=orange!12},
    back/.style={box, fill=green!12},
    dec/.style={box, fill=violet!12},
    lossbox/.style={box, fill=red!10, text width=22mm},
    readout/.style={smallbox, fill=gray!10},
    note/.style={align=center, font=\normalsize\bfseries},
    note2/.style={align=center, font=\small\bfseries},
    arrow/.style={->, thick},
    dashedarrow/.style={->, thick, dashed},
    groupbox/.style={draw, rounded corners, dashed, inner sep=7pt}
]

\node[note] at (10.3,6.2) {Self-supervised training};

\node[proc] (frame) at (0,2.5) {Echo\\frame};

\node[aug] (aug1) at (3.1,3.5) {View 1\\speckle / shadow / contrast / affine};
\node[aug] (aug2) at (3.1,1.0) {View 2\\speckle / shadow / contrast / elastic};

\draw[arrow] (frame.east) -- (aug1.west);
\draw[arrow] (frame.east) -- (aug2.west);

\node[back] (dino1) at (6.7,3.5) {Frozen\\DINOv3\\ViT-S/16};
\node[back] (dino2) at (6.7,1.3) {Frozen\\DINOv3\\ViT-S/16};

\draw[arrow] (aug1.east) -- (dino1.west);
\draw[arrow] (aug2.east) -- (dino2.west);

\node[dec] (adapt1) at (10.0,3.5) {Trainable\\adapter};
\node[dec] (adapt2) at (10.0,1.3) {Trainable\\adapter};

\draw[arrow] (dino1.east) -- (adapt1.west);
\draw[arrow] (dino2.east) -- (adapt2.west);

\node[dec] (dec1) at (13.4,3.5) {U-Net\\decoder};
\node[dec] (dec2) at (13.4,1.3) {U-Net\\decoder};

\draw[arrow] (adapt1.east) -- (dec1.west);
\draw[arrow] (adapt2.east) -- (dec2.west);

\node[procs] (repr1) at (16.9,3.5) {Late dense\\representation};
\node[procs] (repr2) at (16.9,1.3) {Late dense\\representation};

\draw[arrow] (dec1.east) -- (repr1.west);
\draw[arrow] (dec2.east) -- (repr2.west);

\node[lossbox] (byol) at (20.5,2.4) {BYOL\\consistency\\loss};

\draw[arrow] (repr1.east) -- (byol.north west);
\draw[arrow] (repr2.east) -- (byol.south west);

\node[groupbox, fit=(aug1)(aug2)] {};
\node[groupbox, fit=(dino1)(dino2)] {};
\node[groupbox, fit=(adapt1)(adapt2)(dec1)(dec2)(repr1)(repr2)] {};

\node[note2] at (3.1,5.3) {Augmentations};
\node[note2] at (6.5,5.3) {Frozen backbone};
\node[note2] at (13.4,5.3) {Dense adaptation};

\node[note] at (10.3,-0.8) {Downstream evaluation};

\node[readout] (frozen) at (4.0,-2.4) {Frozen hybrid\\SSL model};

\node[readout] (heur) at (8.2,-2.4) {Heuristic\\extraction};
\node[readout] (lin)  at (11.6,-2.4) {Linear\\probe};
\node[readout] (ldec) at (15.0,-2.4) {Lightweight\\decoder};
\node[readout] (pft)  at (18.8,-2.4) {Partial\\fine-tuning};

\node[groupbox, fit=(frozen)(heur)(lin)(ldec)(pft)] (down) {};

\draw[dashedarrow] (byol.south) |- (down.east);
\draw[arrow] (frozen.east) -- (heur.west);
\draw[arrow] (heur.east) -- (lin.west);
\draw[arrow] (lin.east) -- (ldec.west);
\draw[arrow] (ldec.east) -- (pft.west);

\node[align=center, font=\footnotesize] at (11.5,-3.7)
{Evaluation hierarchy: heuristic $\rightarrow$ linear $\rightarrow$ decoder $\rightarrow$ partial fine-tuning};

\end{tikzpicture}%
}
\caption{Schematic of the \ac{ac:BYOS} hybrid \ac{ac:SSL} pipeline. Two augmented views of the same echocardiographic frame are processed by a frozen DINOv3 ViT-S/16 backbone, a trainable adapter, and a U-Net-style decoder. A BYOL-style consistency objective is applied to a late dense representation to encourage spatially coherent and augmentation-invariant anatomical features. The learned frozen representation is then evaluated using heuristic extraction, linear probes, lightweight decoder probes, and partial \ac{ac:FT}.}
\label{fig:byos_pipeline}
\end{figure*}

The augmentation pipeline was tailored to echocardiography and included histogram equalization, median filtering, multiplicative speckle perturbation, simulated shadow bands, small affine perturbations, and mild elastic deformation. These transformations were chosen to encourage invariance to common ultrasound artifacts while preserving anatomy-relevant structure. \ac{ac:BYOS} was therefore not trained to predict \ac{ac:LV} masks directly, but to learn dense and augmentation-stable anatomical structure from unlabeled echocardiographic frames. Figure~\ref{fig:byos_pipeline} summarizes the training and downstream transfer pipeline.

\subsubsection{Hyperparameter search and model selection}
BYOS models were trained on unlabeled EchoNet-Dynamic video frames at $512 \times 512$ resolution using the frozen DINOv3 backbone and a trainable adapter--decoder stack. Because no segmentation labels were used during pretraining, checkpoint selection could not rely on downstream Dice or EF. We therefore performed a small unsupervised hyperparameter search over the dimensionality of the dense decoder output, the learning rate, and the target-encoder momentum decay. In particular, we searched over output dimensionalities from 2 to 7 channels, learning rates including $10^{-3}$, $2 \times 10^{-3}$, and $3 \times 10^{-3}$, and momentum decays spanning 0.5 to 0.99.

Candidate runs were ranked using the \emph{epiplexity} criterion \cite{lit:epiplexity}, together with training-loss behaviour and qualitative preview inspection of the learned dense outputs. Epiplexity was used as an unsupervised proxy for representation richness and non-collapse during training. The final BYOS checkpoint selected for downstream analysis was the five-channel model trained with learning rate $10^{-3}$ and momentum decay $0.95$.

\subsubsection{Interpretation of learned dense channels}

\begin{figure}[!h]
\centering
\includegraphics[width=\linewidth]{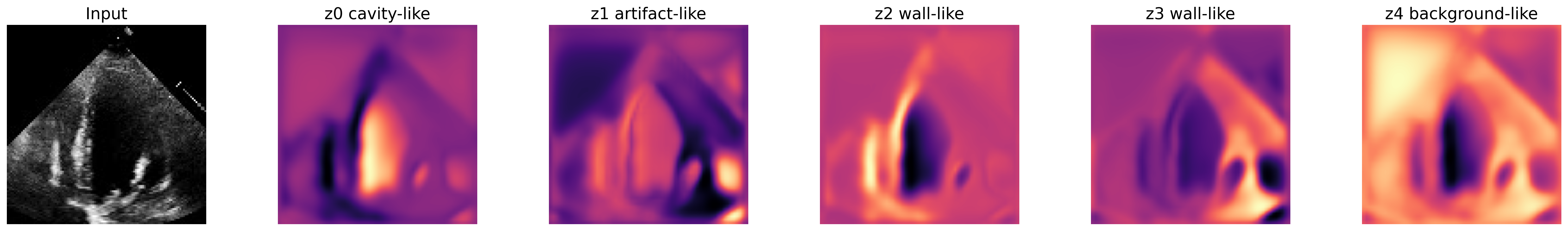}
\caption{Representative dense channel activations from the selected five-channel BYOS model. The learned representation appears to separate broad semantic structure rather than producing an LV-specific mask directly. In the example shown, $z_0$ qualitatively resembles a cavity-like map, $z_1$ an artifact-like map, $z_2$ and $z_3$ complementary wall-like maps, and $z_4$ a background-like map.}
\label{fig:byos_channels}
\end{figure}

After selecting the final BYOS checkpoint, we inspected the dense output channels qualitatively to determine what anatomical or appearance factors they appeared to capture. The learned representation did not produce an LV-specific mask directly. Instead, it appeared to decompose the frame into broader semantic patterns. In the selected five-channel model, one channel qualitatively resembled a \emph{cavity-like} map, highlighting blood-pool regions across multiple chambers. A second channel qualitatively resembled an \emph{artifact-like} map, often activating in peripheral low-information regions and shadowed areas. Two channels qualitatively resembled complementary \emph{wall-like} maps, roughly separating structures on the left and right sides of the image. The final channel appeared predominantly \emph{background-like}. Figure~\ref{fig:byos_channels} shows representative channel activations.

This channel structure explains why direct heuristic extraction from BYOS can recover cavity-like anatomy but still struggles to isolate the LV specifically. The model learns a dense anatomical decomposition rather than an explicit chamber mask. This observation motivated a structured LV extraction stage based on cavity scoring, wall-informed splitting, and temporal tracking.

\subsubsection{Heuristic LV extraction from BYOS predictions}
To test whether the BYOS representation could support chamber recovery without mask-supervised decoding, we constructed a heuristic LV extraction pipeline based on the learned dense channels. First, the cavity-like channel was converted into a per-pixel cavity score, optionally penalized by wall-like, artifact-like, and background-like channels. Candidate cavity masks were then obtained by thresholding the score map, restricting predictions to the ultrasound cone, and applying morphological cleanup.

Because the cavity-like channel often responded to multiple chambers simultaneously, direct thresholding frequently produced connected cavity regions rather than an isolated LV mask. We therefore introduced a wall-informed watershed stage to split connected cavity candidates into chamber-like subregions. The watershed elevation combined wall probability and a neck prior derived from the distance transform, producing candidate parts within each connected cavity blob. These parts were then scored using simple LV-likeness criteria, including approximate location within the apical four-chamber view, size, shape, solidity, border contact, and mean cavity score.

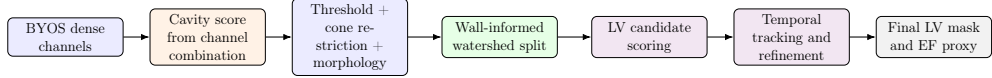
\begin{figure*}[!h]
\centering
\resizebox{0.95\textwidth}{!}{%
\begin{tikzpicture}[
    font=\small,
    >=Latex,
    box/.style={draw, rounded corners, align=center, minimum height=10mm, text width=28mm, inner sep=4pt},
    proc/.style={box, fill=blue!8},
    score/.style={box, fill=orange!10},
    split/.style={box, fill=green!10},
    select/.style={box, fill=violet!10},
    finalbox/.style={box, fill=gray!10},
    arrow/.style={->, thick}
]
\node[proc]   (in)    at (0,0)   {BYOS dense\\channels};
\node[score]  (score) at (3.8,0) {Cavity score\\from channel combination};
\node[proc]   (thr)   at (7.6,0) {Threshold +\\cone restriction +\\morphology};
\node[split]  (ws)    at (11.6,0) {Wall-informed\\watershed split};
\node[select] (cand)  at (15.6,0) {LV candidate\\scoring};
\node[select] (track) at (19.4,0) {Temporal\\tracking and\\refinement};
\node[finalbox]    (mask)  at (23.2,0) {Final LV mask\\and EF proxy};

\draw[arrow] (in.east) -- (score.west);
\draw[arrow] (score.east) -- (thr.west);
\draw[arrow] (thr.east) -- (ws.west);
\draw[arrow] (ws.east) -- (cand.west);
\draw[arrow] (cand.east) -- (track.west);
\draw[arrow] (track.east) -- (mask.west);
\end{tikzpicture}%
}
\caption{Schematic of heuristic LV extraction from the BYOS representation. Dense channels are converted into a cavity score, thresholded and constrained to the ultrasound cone, split into chamber-like parts using wall-informed watershed, scored for LV-likeness, and linked across time to produce the final LV mask sequence.}
\label{fig:byos_lv_extraction}
\end{figure*}

\begin{figure}[!h]
\centering
\includegraphics[width=\linewidth]{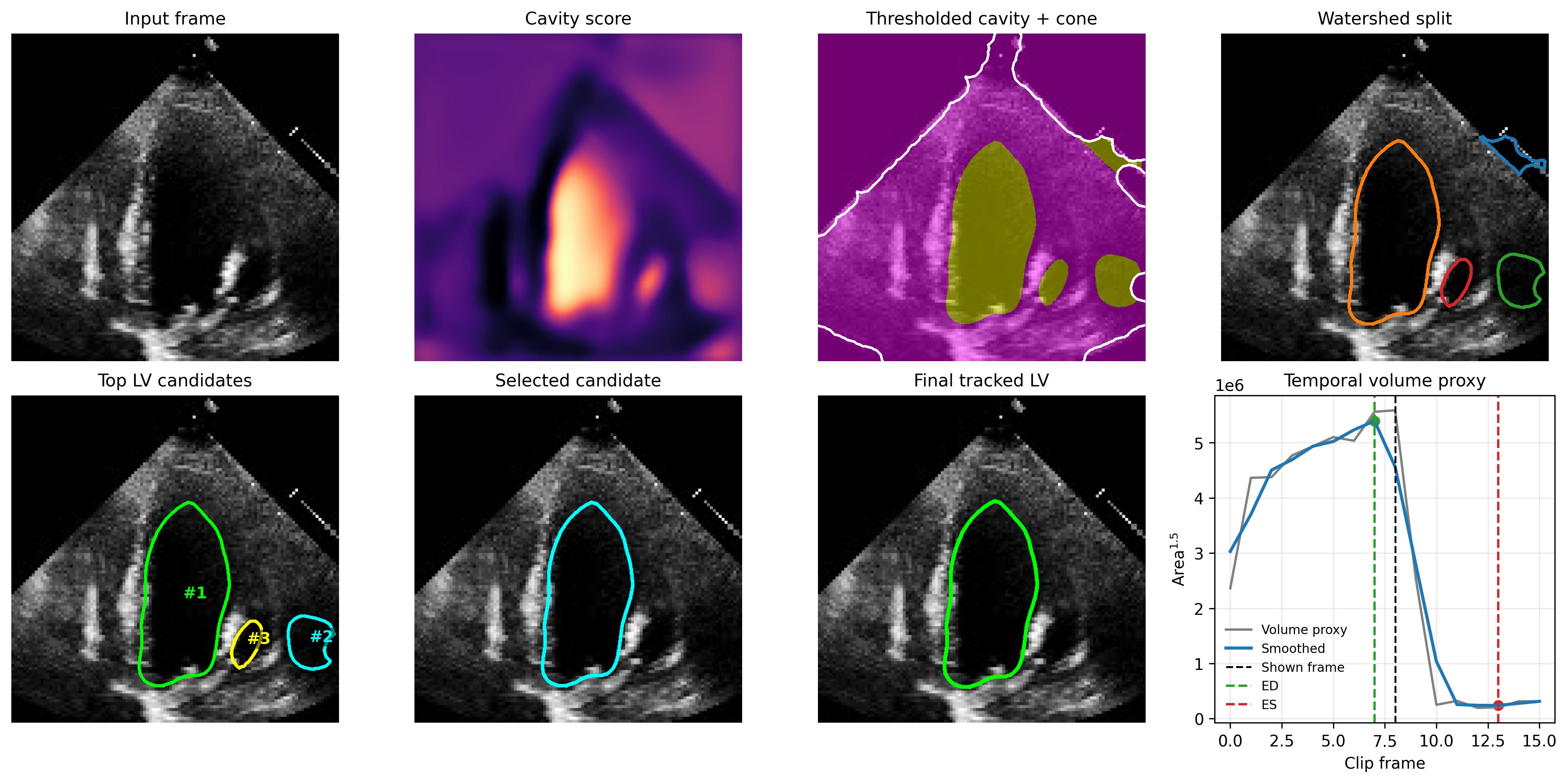}
\caption{Representative BYOS-based heuristic LV extraction pipeline for a single frame within a short video clip. The cavity-like dense channel is converted into a cavity score, thresholded and restricted to the ultrasound cone, and then split into chamber-like regions using wall-informed watershed. Candidate LV regions are scored by simple anatomical priors, after which a temporally regularized path selects the final LV mask sequence. The final panel shows the resulting temporal volume proxy used for ED/ES selection and EF estimation.}
\label{fig:byos_lv_extraction_vis}
\end{figure}

For video-level inference, candidate LV regions were linked across time using a temporally regularized beam-search procedure. The resulting path was further refined by local temporal consistency checks to suppress abrupt centroid shifts, implausible area jumps, and unstable shape changes. Hyperparameters for cavity scoring, thresholding, watershed splitting, candidate pruning, and temporal tracking were tuned on a subset of validation videos by balancing mean Dice on labeled frames against EF error. The final selected configuration used fixed thresholding, cone masking, wall-informed watershed separation, and anchor-based temporal tracking. Figures~\ref{fig:byos_lv_extraction}-\ref{fig:byos_lv_extraction_vis} summarize this heuristic LV extraction pipeline.

This heuristic stage should therefore be interpreted not as a learned segmentation head, but as a structured extraction strategy used to test whether the BYOS representation already contains enough information to support LV recovery without mask-supervised training.

\subsection{DINOv3 heuristic extraction}
\label{subsec:dino_heuristic}

To construct a no-mask-training baseline using a generic foundation model, we applied heuristic LV extraction to frozen DINOv3 patch-token features. This baseline used a frozen DINOv3 ViT-S/16 encoder and did not train any segmentation head or EF predictor. Instead, each frame was converted into a patch-level score map intended to highlight cavity-like regions, which was then transformed into a binary mask by fixed hand-designed rules.

The purpose of this baseline was to test whether a strong generic self-supervised representation is sufficient, by itself, to support LV extraction without any supervised extraction strategy. In this setting, the DINOv3 backbone provides frozen patch-level spatial features, but the mapping from features to LV masks is entirely heuristic. The method should therefore be interpreted as a \emph{feature-guided cavity extractor} rather than a learned segmentation model.

For each video, frames were resized to a fixed spatial resolution and passed through the frozen DINOv3 encoder. Patch tokens were arranged on their native spatial grid and used as the structural support for frame-level scoring. In practice, the dominant cue in this heuristic baseline was patchwise darkness rather than an explicitly learned chamber score. This choice was motivated by the observation that the LV cavity is often relatively hypoechoic compared with surrounding myocardium, although this is only a weak and non-specific prior.

For each frame $t$, a patchwise darkness map $D_t$ was computed from the grayscale image so that darker regions corresponded to higher scores. Two optional priors were added. First, a spatial center prior $C$ favored regions closer to the central portion of the ultrasound sector, reducing selection of peripheral dark regions near the cone boundary. Second, a temporal-variance prior $V$ was computed across the video under the assumption that functionally relevant intracardiac regions may show greater temporal variation than static background structures. The final score map was defined as
\begin{equation}
S_t = w_d D_t + w_c C + w_v V,
\end{equation}
where $w_d$, $w_c$, and $w_v$ denote the darkness, center-prior, and temporal-variance weights.

The patch-level score map was normalized independently for each frame and then upsampled to image resolution using bilinear interpolation. A binary candidate mask was obtained by thresholding the resulting score image, restricting it to the ultrasound cone, and applying standard morphological cleanup, including removal of small connected components, closing, optional opening, and hole filling. Depending on the configuration, the final mask was taken either as the full postprocessed region or as its largest connected component.

This heuristic baseline did not involve backpropagation or mask-supervised optimization. Instead, a small validation sweep was performed over the threshold, center-prior weight, temporal-variance weight, and minimum connected-component size to identify a reasonable fixed rule. The selected configuration was then evaluated as a frozen, chamber-unspecific heuristic baseline.

\subsection{Evaluation of heuristic baselines}
\label{subsec:heuristic_eval}

The heuristic baselines were evaluated using the same segmentation and functional protocol as the learned downstream methods. On the traced frames provided by EchoNet-Dynamic, predicted masks were compared with the corresponding ground-truth \ac{ac:LV} masks using Dice and \ac{ac:IoU}. Across sampled video frames, we also computed temporal stability statistics, including temporal \ac{ac:IoU}, area coefficient of variation, border-touch fraction, centroid jitter, and a simple cavity-to-surround contrast proxy.

To obtain a video-level functional estimate, the full sequence of predicted masks was converted into a surrogate volume curve using the same area-based \ac{ac:EF} proxy as all other segmentation-derived methods. Candidate end-diastolic and end-systolic frames were selected from this curve, and \ac{ac:EF} was computed from the resulting surrogate volumes. We report video-level \ac{ac:EF} \ac{ac:MAE} against the reference \ac{ac:EF} labels, as well as a two-frame oracle variant using traced frames when available.

These heuristic baselines therefore address a specific methodological question: whether a frozen self-supervised representation can support clinically meaningful LV extraction without any supervised extraction strategy. These baselines were intended as restricted no-mask-training extraction regimes for probing representational accessibility, not as optimized label-free segmentation methods. Although no segmentation model was trained for these baselines, validation annotations were used to select heuristic hyperparameters before final test evaluation. Weak performance in this setting should not be interpreted as evidence that the representation itself is weak; rather, it may indicate that the representation is being under-read by a chamber-unspecific heuristic rule. This distinction motivates the subsequent comparison with frozen linear probes, lightweight decoder probes, and partial \ac{ac:FT}.

\subsection{Frozen supervised probes}
\label{subsec:frozen_probes}

To determine how much \ac{ac:LV} information is recoverable from a frozen representation, we trained supervised probe heads using traced frames from the training split, selected model configurations using the validation split, and report final results on the test split. Two probe types were used.

A \emph{linear probe} consisted of a minimal pointwise projection from frozen features to mask logits. This probe tests whether \ac{ac:LV} structure is directly and approximately linearly separable in the representation.

A \emph{lightweight decoder probe} consisted of a compact convolutional upsampling decoder with hidden width 128, bilinear upsampling between stages, and a final $1 \times 1$ output layer. This probe retains limited nonlinear and local spatial refinement capacity while keeping the backbone frozen. It tests whether the frozen representation contains richer LV information that is not fully linearly separable but can be recovered by a small nonlinear extraction strategy. Exact probe configurations are provided in the Supplementary Material.

For the BYOS representation, the frozen feature source was taken from a late intermediate spatial layer. For DINOv3, patch-token grids were used as the frozen feature input to the supervised probe.

\subsection{Partial fine-tuning}
\label{subsec:partial_ft}

To test whether limited supervised adaptation improves over frozen-probe transfer, we partially fine-tuned both the BYOS and DINOv3 representation families. In each case, a lightweight supervised decoder was trained jointly with a restricted subset of backbone parameters while the remainder of the representation was kept frozen. These experiments test whether modest task-specific adaptation provides a meaningful gain over strong frozen-feature baselines, or whether the frozen representation already contains most of the recoverable LV information.

For BYOS, a lightweight decoder head was attached at the feature map produced by \texttt{cnn.encoder.adapt}. All model parameters were frozen initially, after which only selected modules were unfrozen. Two configurations were studied: unfreezing only \texttt{cnn.encoder.adapt}, and unfreezing both \texttt{cnn.encoder.adapt} and \texttt{cnn.decoder.blocks.3}.

For DINOv3, patch-token features from the ViT-S/16 backbone were decoded using the same lightweight decoder family as in the frozen decoder probe. We then evaluated two restricted unfreezing configurations: unfreezing the final transformer block together with the final normalization layer, and unfreezing the final two transformer blocks together with the final normalization layer. In all cases, the decoder head and the unfrozen backbone modules used separate learning rates, with the backbone updated more conservatively.

This setup was chosen to preserve the pretrained representation as much as possible while allowing limited task-specific adaptation. By applying analogous partial \ac{ac:FT} analyses to both representation families, we aimed to assess whether the main gains arise from increased extraction strategy capacity alone or from additional supervised adaptation of the pretrained backbone.

\subsection{Supervised baselines}

We compared the self-supervised methods against two supervised references. The first was a standard U-Net-style segmentation model trained directly on the traced frame masks. This baseline provides the main full-supervision reference for segmentation-derived \ac{ac:EF}. The second was a direct \ac{ac:EF} regressor trained to predict video-level \ac{ac:EF} without explicitly producing segmentation masks. This baseline tests whether explicit segmentation is necessary for best functional performance.

\subsection{Segmentation training objective}

For mask-supervised segmentation heads, including linear probes, lightweight decoder probes, partial \ac{ac:FT}, and the supervised U-Net, we optimized a combination of binary cross-entropy and soft Dice loss:
\begin{equation}
\mathcal{L}_{\mathrm{seg}}
=
\lambda_{\mathrm{BCE}} \mathcal{L}_{\mathrm{BCE}}
+
\lambda_{\mathrm{Dice}} \mathcal{L}_{\mathrm{Dice}}.
\end{equation}

The soft Dice term was defined as
\begin{equation}
\mathcal{L}_{\mathrm{Dice}}
=
1 -
\frac{2 \sum_i p_i y_i + \epsilon}
{\sum_i p_i + \sum_i y_i + \epsilon},
\end{equation}
where $p_i$ denotes the predicted probability at pixel $i$ and $y_i$ the corresponding binary target. In practice, equal weighting between the \ac{ac:BCE} and Dice terms was used.

\subsection{Mask postprocessing and temporal stability analysis}

At inference time, per-frame logits were converted into probabilities using a sigmoid and thresholded to form binary masks. Postprocessing included removal of small connected components, morphological closing, optional opening, hole filling, and optional retention of the largest connected component. When enabled, the final mask was restricted to the ultrasound cone region.

To better characterize temporal behavior, we also computed auxiliary stability metrics on sampled frames from each video. These included temporal IoU between adjacent predicted masks, area coefficient of variation, centroid jitter, border-touch fraction, and a simple contrast proxy comparing cavity intensity to a surrounding ring. These measures were not used for model optimization, but helped interpret failure modes such as unstable cavity size, peripheral leakage, and excessive border contact.

\subsection{EF proxy from predicted masks}
\label{subsec:ef_proxy}

To obtain \ac{ac:EF} from a sequence of predicted masks, we first measured the predicted cavity area for every frame. Let $A_t$ denote the binary LV area at frame $t$. We then used a simple area-to-volume surrogate:
\begin{equation}
V_t = A_t^{\gamma},
\end{equation}
with $\gamma = 1.5$ in our experiments. This exponent was fixed across all segmentation-derived methods and used as a common surrogate rather than as a calibrated volumetric model. The resulting volume sequence was smoothed temporally, and candidate end-diastolic and end-systolic frames were selected from the largest and smallest values, respectively, subject to a minimum temporal separation. EF was then computed as
\begin{equation}
\mathrm{EF}
=
100 \times
\frac{V_{\mathrm{ED}} - V_{\mathrm{ES}}}{V_{\mathrm{ED}}}.
\end{equation}

This formulation provided a simple and consistent functional proxy across segmentation-derived methods. To better separate segmentation quality from automatic frame-selection error, we also computed a two-frame oracle variant using traced frames when available. This provided a more favorable estimate of segmentation-derived EF when ED/ES ambiguity was reduced.

\subsection{Evaluation metrics}

Segmentation accuracy on labeled frames was measured using Dice and IoU:
\begin{equation}
\mathrm{Dice}(P,G)
=
\frac{2 |P \cap G|}{|P| + |G|},
\end{equation}
\begin{equation}
\mathrm{IoU}(P,G)
=
\frac{|P \cap G|}{|P \cup G|},
\end{equation}
where $P$ is the predicted mask and $G$ the ground-truth mask.

Functional accuracy was measured using mean absolute error between predicted EF and reference EF:
\begin{equation}
\mathrm{MAE}
=
\frac{1}{N}
\sum_{i=1}^{N}
|\hat{e}_i - e_i|,
\end{equation}
where $\hat{e}_i$ is the predicted EF for video $i$ and $e_i$ is the corresponding reference value. We additionally report RMSE and median absolute error in the experiment summaries, but EF MAE was the primary functional endpoint.

\subsection{Statistical uncertainty and paired bootstrap analysis}
\label{subsec:bootstrap_stats}

To quantify uncertainty in downstream performance estimates, we assessed statistical variability at the \emph{video level} using nonparametric bootstrap resampling on the EchoNet-Dynamic test split. Video-level resampling was chosen because EF is defined per video and because segmentation results were summarized as one mean overlap score per video across the available labeled frames. This avoids overstating confidence by treating multiple frames from the same video as independent observations.

For each method, we computed 95\% bootstrap confidence intervals for Dice, IoU, and EF mean absolute error (MAE) using 5000 bootstrap replicates over test videos. Dice and IoU were computed from the per-video mean overlap on labeled frames, and EF MAE was computed from the absolute difference between the predicted video-level EF and the reference EF.

For key pairwise method comparisons, we used \emph{paired} bootstrap resampling over the common set of test videos evaluated by both methods. In each replicate, the same resampled set of videos was used for both methods, and the mean difference in the metric of interest was computed. We report these paired mean differences together with 95\% bootstrap confidence intervals. For error metrics such as EF MAE, a negative paired difference indicates lower error for the second method in the comparison. This paired design preserves within-video correspondence across methods and provides a more informative assessment of performance differences than unpaired comparisons.

\subsection{Implementation details}

All models were implemented in PyTorch. Detailed implementation settings for each method family are summarized in Supplementary Section A. 
Unless otherwise stated, segmentation-derived methods used $512 \times 512$ inputs and AdamW optimization. Frozen-probe and partial fine-tuning experiments were trained on traced EchoNet-Dynamic frames using an equal-weight combination of binary cross-entropy and soft Dice loss. For these mask-supervised segmentation models, training augmentation consisted of random horizontal flipping with probability 0.5 and mild brightness/contrast perturbation with magnitude $\pm 0.10$.

During inference, segmentation-derived models processed videos in temporal chunks of 32 frames to reduce memory consumption. Binary masks were obtained by thresholding sigmoid probabilities at 0.5, followed by removal of connected components smaller than 300 pixels, morphological closing with radius 3, optional opening, hole filling, retention of the largest connected component when enabled, and restriction to the inferred ultrasound cone when enabled. Unless otherwise stated, models were selected using validation-set performance, with validation Dice used for segmentation models and validation mean absolute error used for direct EF regression.

\subsection{Experimental hypothesis}

The central hypothesis of this study was that self-supervised echocardiographic representations contain substantially more LV structure than is suggested by heuristic extraction alone. If this hypothesis is correct, performance should improve systematically as the extraction strategy becomes more expressive: heuristic extraction should be weakest, linear probes should recover more LV information, lightweight decoder probes should approach supervised segmentation quality, and partial \ac{ac:FT} should reveal whether meaningful gains remain available through limited adaptation of the pretrained representation.

\section{Results}
\label{sec:results}

\subsection{Overall comparison across methods}

Table~\ref{tab:main_results} summarizes the main quantitative results on the EchoNet-Dynamic test set. Among the segmentation-based methods, the supervised U-Net achieved the highest overlap, with Dice $0.915$ and IoU $0.848$. Among the self-supervised transfer settings, the strongest frozen DINOv3 decoder probe achieved Dice $0.906$, IoU $0.834$, and EF MAE $9.65$, while the frozen BYOS decoder probe achieved Dice $0.902$, IoU $0.827$, and the lowest segmentation-derived EF MAE in the study at $8.74$. These results indicate that both a generic frozen foundation representation and a task-adapted dense self-supervised representation can support strong downstream performance when paired with an appropriate decoder.

In contrast, heuristic extraction was markedly weaker. The BYOS heuristic baseline achieved Dice $0.687$, IoU $0.558$, and EF MAE $17.83$, while the DINOv3 heuristic baseline achieved Dice $0.684$, IoU $0.537$, and EF MAE $13.01$. For both representation families, performance improved substantially when moving from heuristic extraction to learned downstream probes, and improved further when moving from linear probes to lightweight decoder probes. This pattern is consistent with the central premise of the study: the apparent utility of a frozen representation depends strongly on the downstream extraction strategy used to evaluate it.

Partial \ac{ac:FT} yielded comparatively modest changes relative to the strongest frozen-decoder baselines. For DINOv3, unfreezing the final transformer block together with the final normalization layer increased overlap to Dice $0.910$ and IoU $0.840$, but EF MAE was $10.26$. Unfreezing the final two transformer blocks together with the final normalization layer yielded Dice $0.911$, IoU $0.841$, and EF MAE $9.57$, numerically close to the frozen decoder baseline. For BYOS, both partial \ac{ac:FT} configurations likewise produced only modest changes relative to the frozen decoder probe. Overall, these results suggest that the main gains in this setting arise from improving the downstream extraction strategy rather than from limited supervised adaptation of the pretrained representation.

The direct EF regressor achieved the lowest learned functional error overall, with EF MAE $6.65$, while the ground-truth-mask EF proxy ceiling achieved EF MAE $3.33$ under the same proxy formulation. This indicates that \linebreak segmentation-derived EF remains constrained by segmentation quality, chamber specificity, and temporal frame selection even when overlap is high.

\begin{table}[!h]
\centering
\scriptsize
\setlength{\tabcolsep}{2.5pt}
\renewcommand{\arraystretch}{1.04}
\caption{Main quantitative comparison on the EchoNet-Dynamic test set. Dice and IoU are computed on labeled frames and summarized at the video level. EF metrics are computed from segmentation-derived EF unless otherwise noted. Best values within each section are shown in bold. The marker (†) indicates that the EF MAE difference was not clearly distinguishable from the relevant strong frozen baseline under paired bootstrap analysis. Specifically, among DINOv3-based methods, the frozen decoder and the partially fine-tuned model with the last two blocks unfrozen were not clearly separable in EF MAE; likewise, the supervised U-Net was not clearly separable in EF MAE from the frozen DINOv3 decoder. Corresponding 95\% video-level bootstrap confidence intervals and paired bootstrap comparisons are reported in the Supplementary Material.}
\label{tab:main_results}
\begin{tabularx}{\linewidth}{
>{\raggedright\arraybackslash}p{3.75cm}
>{\raggedright\arraybackslash}p{1.85cm}
>{\raggedright\arraybackslash}X
>{\centering\arraybackslash}p{0.78cm}
>{\centering\arraybackslash}p{0.78cm}
>{\centering\arraybackslash}p{0.90cm}
>{\centering\arraybackslash}p{0.90cm}}
\toprule
\textbf{Method} & \textbf{Backbone} & \textbf{Trainable components} & \textbf{Dice} & \textbf{IoU} & \textbf{MAE} & \textbf{RMSE} \\
\midrule
\multicolumn{7}{l}{\textit{Segmentation-derived methods}} \\
BYOS heuristic & BYOS SSL & none & 0.687 & 0.558 & 17.83 & 22.62 \\
DINOv3 heuristic & DINOv3 ViT & none & 0.684 & 0.537 & 13.01 & 16.34 \\
BYOS + linear probe & BYOS SSL & linear head & 0.838 & 0.730 & 13.49 & 17.05 \\
DINOv3 + linear probe & DINOv3 ViT & linear head & 0.853 & 0.752 & 14.07 & 17.50 \\
BYOS + decoder & BYOS SSL & decoder & 0.902 & 0.827 & \best{8.74} & \best{11.67} \\
DINOv3 + decoder & DINOv3 ViT & decoder & 0.906 & 0.834 & 9.65 (†) & 12.59 \\
Partial FT BYOS + decoder (adapt only) & BYOS SSL & decoder + adapt & 0.906 & 0.833 & 9.71 & 12.66 \\
Partial FT BYOS + decoder (adapt + dec3) & BYOS SSL & decoder + sel. blocks & 0.906 & 0.833 & 9.27 & 12.35 \\
Partial FT DINOv3 + decoder (last block + norm) & DINOv3 ViT & decoder + last block + norm & 0.910 & 0.840 & 10.26 & 13.06 \\
Partial FT DINOv3 + decoder (last 2 blocks + norm) & DINOv3 ViT & decoder + last 2 blocks + norm & 0.911 & 0.841 & 9.57 (†) & 12.62 \\
Supervised U-Net & CNN U-Net & full model & \best{0.915} & \best{0.848} & 9.72 (†) & 12.58 \\
\midrule
\multicolumn{7}{l}{\textit{Functional reference baselines}} \\
Direct EF regressor & CNN / video & full model & -- & -- & 6.65 & 9.20 \\
GT-mask EF proxy ceiling & N/A & none & 1.000 & 1.000 & \best{3.33} & \best{4.33} \\
\bottomrule
\end{tabularx}
\end{table}

Video-level bootstrap analysis supported the stability of the main point estimates. For example, the frozen DINOv3 decoder achieved Dice $0.906$ (95\% CI $0.903$--$0.909$) and EF MAE $9.65$ (95\% CI $9.20$--$10.09$), the partially fine-tuned BYOS configuration (adapt only) achieved Dice $0.906$ (95\% CI $0.903$--$0.909$) and EF MAE $9.71$ (95\% CI $9.26$--$10.15$), and the supervised U-Net achieved Dice $0.915$ (95\% CI $0.913$--$0.918$) and EF MAE $9.72$ (95\% CI $9.29$--$10.16$). Confidence intervals and paired bootstrap comparisons for the main method families are reported in Supplementary Section B.
Among the strongest segmentation-derived methods, overlap differences were somewhat clearer than functional differences. In particular, the frozen DINOv3 decoder and the partially fine-tuned DINOv3 model with the final two transformer blocks unfrozen were separated slightly in overlap, but not clearly in EF MAE under paired bootstrap analysis.

\subsection{Heuristic extraction underestimates recoverable representation utility}

A consistent result across both representation families was that heuristic extraction provided a substantially weaker estimate of downstream utility than learned probes. For DINOv3, the heuristic baseline achieved Dice $0.684$ and EF MAE $13.01$, while the frozen linear probe improved overlap to Dice $0.853$. The frozen lightweight decoder probe improved further to Dice $0.906$ and EF MAE $9.65$. Partial \ac{ac:FT} produced only small additional changes relative to this frozen-decoder baseline. These results indicate that the frozen DINOv3 representation contains substantially more \ac{ac:LV}-relevant information than is accessible through the heuristic extraction rule alone.

A similar pattern was observed for BYOS. The heuristic baseline achieved Dice $0.687$ and EF MAE $17.83$, whereas the frozen linear probe improved performance to Dice $0.838$ and EF MAE $13.49$. The frozen lightweight decoder probe improved further to Dice $0.902$ and EF MAE $8.74$. Both partial \ac{ac:FT} configurations again produced only modest changes relative to the strongest frozen-decoder result. Across both families, the largest performance change was therefore not between frozen transfer and partial \ac{ac:FT}, but between restricted heuristic extraction and more expressive frozen downstream decoding.

\begin{figure*}[!h]
\centering
\includegraphics[width=0.88\textwidth]{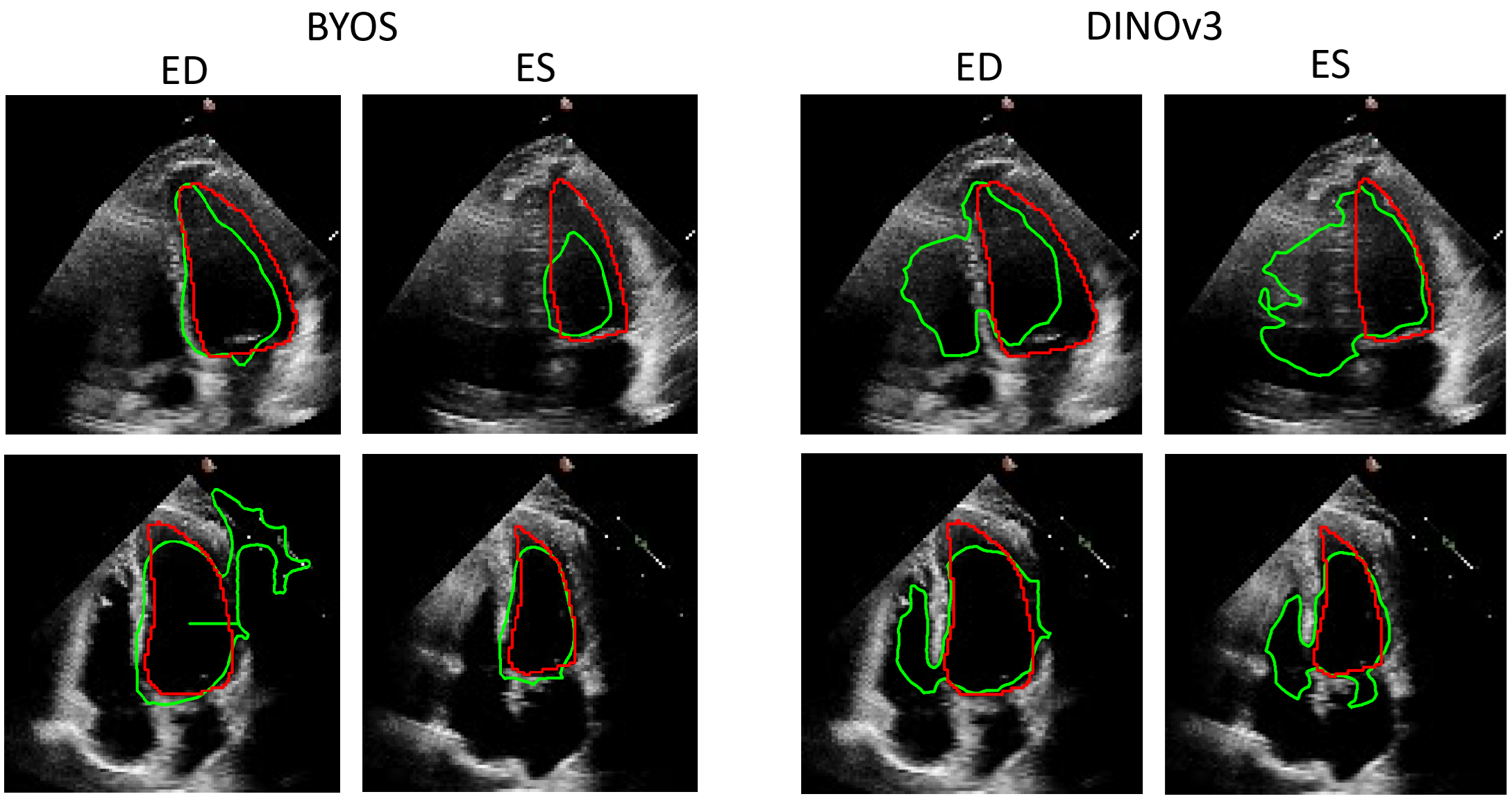}
\caption{Representative qualitative outputs of the heuristic baselines. The left pair of columns shows BYOS-based heuristic extraction and the right pair shows DINOv3-based heuristic extraction; rows correspond to different videos, with ED and ES frames shown for each case. Predicted contours are shown in green and ground-truth LV contours in red. The BYOS heuristic appears more structured because it includes wall-informed chamber separation and temporal selection, whereas the DINOv3 heuristic is shown in its simpler non-watershed form. Despite this difference, both heuristic families remain limited by incomplete LV specificity and tend to include merged chambers, peripheral leakage, or anatomically oversized regions.}
\label{fig:qualitative_failures}
\end{figure*}

These results support a cautious interpretation of weak heuristic performance in dense self-supervised evaluation. In the present setting, poor heuristic results did not imply poor frozen representations; rather, they reflected limited access to information already present in the feature space. Figure~\ref{fig:qualitative_failures} illustrates representative failure modes of the heuristic baselines. Although the BYOS heuristic appears more structured than the DINOv3 heuristic because it incorporates chamber separation and temporal selection, both remained limited by incomplete \ac{ac:LV} specificity. Common failure modes included merged chambers, peripheral leakage, and anatomically oversized masks. The qualitative examples are therefore consistent with the quantitative findings in suggesting that self-supervised representations can encode useful cardiac structure even when weak extraction rules fail to recover it reliably.

\subsection{Frozen decoder probes recover most of the measurable gain}

The frozen supervised probes provided an informative intermediate view between heuristic extraction and partial \ac{ac:FT}. For both DINOv3 and BYOS, linear probes improved substantially over heuristics, indicating that \ac{ac:LV} information was at least partly decodable from the frozen representations. In both families, lightweight decoder probes improved further, suggesting that the frozen features contained richer spatial structure that was not fully accessible to a pointwise linear mapping but could be recovered by limited nonlinear decoding.

Beyond this point, the gains from partial \ac{ac:FT} were modest. For BYOS, the frozen decoder probe achieved the lowest segmentation-derived EF MAE in the study at $8.74$. Unfreezing only \texttt{cnn.encoder.adapt} yielded Dice $0.906$, IoU $0.833$, and EF MAE $9.71$, while additionally unfreezing \linebreak \texttt{cnn.decoder.blocks.3} yielded Dice $0.906$, IoU $0.833$, and EF MAE $9.27$. For DINOv3, relative to the frozen decoder probe (Dice $0.906$, IoU $0.834$, EF MAE $9.65$), unfreezing the final transformer block and normalization layer improved overlap to Dice $0.910$ and IoU $0.840$ but did not improve EF MAE. Unfreezing the final two transformer blocks and normalization layer yielded Dice $0.911$, IoU $0.841$, and EF MAE $9.57$, which was numerically similar to the frozen decoder baseline and not clearly distinguishable from it in paired bootstrap analysis. Overall, these comparisons suggest that in the present setting most of the measurable downstream benefit was already captured by strong frozen-decoder transfer.

\subsection{Segmentation overlap and functional accuracy are related but not equivalent}

Across methods, better segmentation overlap generally corresponded to lower EF error, but the ranking was not strictly monotonic. The supervised U-Net achieved the highest Dice among the learned segmentation models, whereas the frozen BYOS decoder achieved the lowest segmentation-derived EF MAE. Similarly, the strongest frozen and partially fine-tuned DINOv3 variants were close in Dice and IoU and differed only modestly in EF MAE. This indicates that overlap-based segmentation quality and downstream functional accuracy are related but not interchangeable criteria.

\begin{figure*}[!h]
\centering
\includegraphics[width=0.9\linewidth]{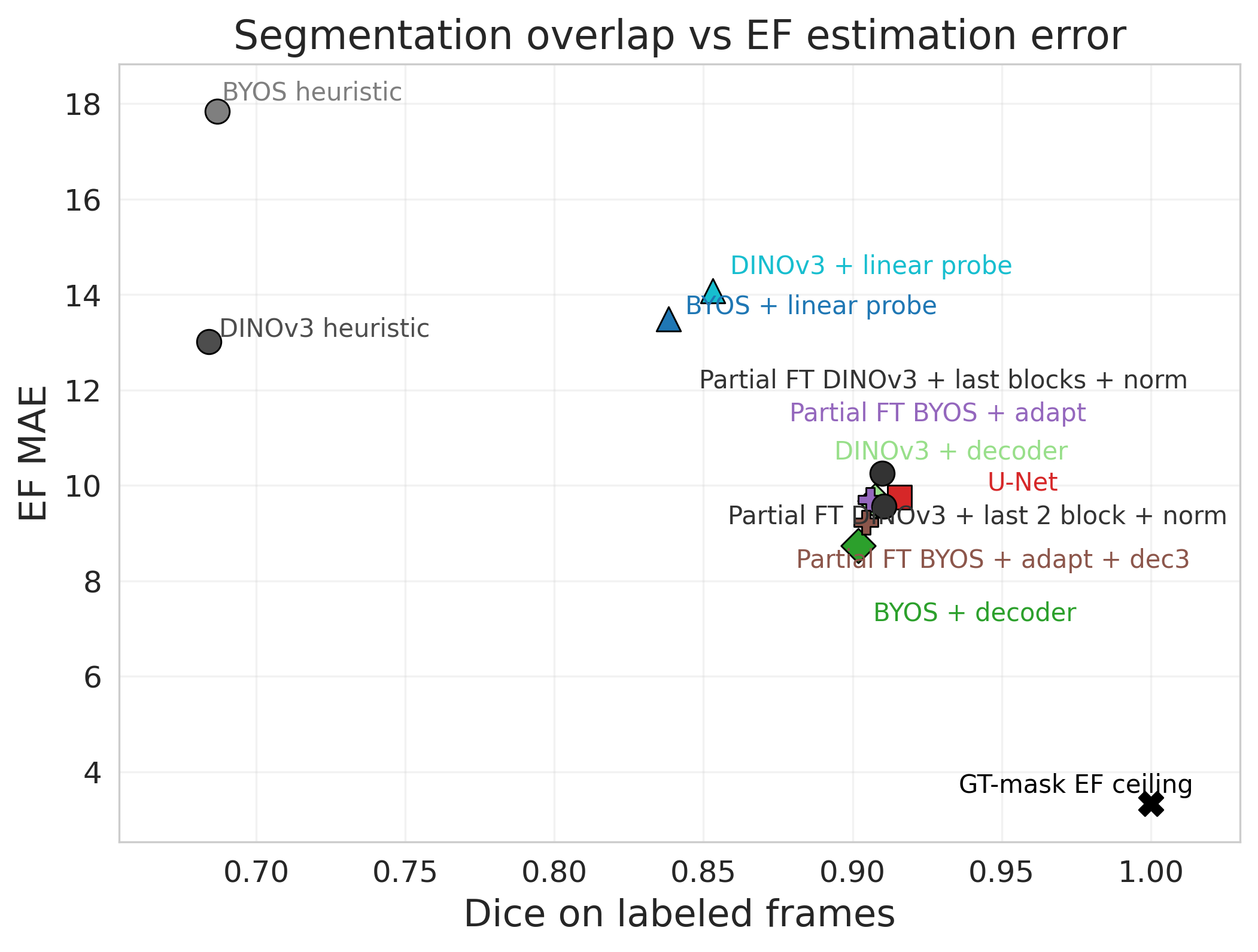}
\caption{Relationship between segmentation overlap and EF estimation accuracy across evaluated methods. Better Dice generally corresponds to lower EF error, but the ranking is not strictly monotonic, indicating that overlap-based segmentation quality and functional performance are related but not equivalent.}
\label{fig:dice_vs_ef}
\end{figure*}

A noteworthy example is provided by the DINOv3 heuristic and linear-probe baselines. The linear probe substantially improved Dice relative to the heuristic baseline, but its EF MAE was slightly higher. This suggests that improved frame-level overlap alone does not guarantee improved functional estimation, likely because segmentation-derived EF is also sensitive to factors such as chamber specificity, area bias, and temporal consistency across the video. More generally, small contour differences or leakage patterns may have limited effect on Dice while still altering the surrogate volume curve used for ED/ES selection and EF estimation. Figure~\ref{fig:dice_vs_ef} visualizes the overall relationship between segmentation overlap and EF performance across methods.

\subsection{Direct EF prediction provides the strongest learned functional baseline}

The direct EF regressor achieved the lowest learned functional error overall, with EF MAE $6.65$, outperforming all segmentation-derived learned methods. This confirms that when the sole endpoint is EF prediction, direct optimization on EF remains highly effective. However, the direct EF baseline does not produce explicit chamber masks and therefore does not provide the same degree of anatomical interpretability as segmentation-based approaches.

The GT-mask EF ceiling achieved EF MAE $3.33$ under the same area-to-volume proxy. This suggests that the proxy itself is relatively strong in this setting and that a substantial component of the remaining error in segmentation-derived pipelines arises upstream, from segmentation quality, chamber specificity, and temporal frame selection rather than from the proxy formulation alone.

Overall, the results support three main conclusions. First, heuristic extraction substantially underestimates the recoverable utility of strong frozen self-supervised representations for dense echocardiographic analysis. Second, lightweight decoder probes recover most of the measurable gain, while limited partial \ac{ac:FT} provides only modest additional benefit. Third, segmentation overlap and functional accuracy are related but not equivalent, so evaluation of segmentation-derived clinical pipelines benefits from considering both anatomical and functional endpoints.

\section{Discussion}
\label{sec:discussion}

This study examined how conclusions about self-supervised representation quality for dense echocardiographic analysis depend on the downstream extraction strategy used for evaluation. Across two complementary representation families, a generic frozen foundation representation based on DINOv3 and a task-adapted dense self-supervised representation (\ac{ac:BYOS}), the same overall pattern was observed. Heuristic extraction provided a weak estimate of what was recoverable from the frozen representation, linear probing revealed substantially more \ac{ac:LV} information, and lightweight decoder probes recovered markedly stronger anatomical and functional performance. By contrast, limited partial \ac{ac:FT} produced only modest additional changes relative to strong frozen-decoder baselines. Taken together, these findings indicate that, for dense echocardiographic prediction, apparent representation quality is shaped not only by the representation itself but also by the downstream mechanism used to access it.

The clearest methodological implication is that evaluation under a single downstream probe can be misleading in dense medical imaging settings. For tasks such as \ac{ac:LV} segmentation, downstream performance reflects a combination of representational content and extraction capacity. A weak result may arise because the representation lacks task-relevant information, because the probe is underpowered, or because the chosen downstream model is poorly matched to the way anatomical structure is encoded in the representation. In the present experiments, both DINOv3 and \ac{ac:BYOS} appeared relatively weak under heuristic extraction, yet both supported strong segmentation and substantially improved segmentation-derived \ac{ac:EF} estimation once paired with a modest nonlinear decoder. This suggests that weak heuristic or minimal-probe performance should be interpreted cautiously when assessing self-supervised representations for dense clinical tasks.

This issue is particularly salient in echocardiography. Compared with many natural-image benchmarks, cardiac ultrasound combines low signal-to-noise ratio, weak and often discontinuous chamber boundaries, substantial acquisition variability, and frequent visual continuity between adjacent structures. Under these conditions, useful anatomical information may be present in a frozen representation without being directly recoverable by a chamber-unspecific heuristic rule or a pointwise linear mapping. The observed progression from heuristic extraction to linear probing and then to lightweight decoder probing is consistent with this interpretation. The results suggest that a substantial portion of the recoverable \ac{ac:LV} structure is already present in strong frozen features, but that accessing it reliably requires limited nonlinear spatial decoding.

The comparison between the two representation families is informative in this respect. DINOv3 provides a generic self-supervised feature space learned without echocardiography-specific supervision, whereas \ac{ac:BYOS} introduces ultrasound-adapted dense self-supervision on top of a frozen DINOv3 backbone. Although the two families differ in construction and in the qualitative structure of their frozen features, both showed the same extraction-strategy-sensitive pattern. This consistency strengthens the central claim of the paper: the dependence of apparent representation quality on downstream extraction strategy is not specific to a single pretrained feature source or architectural design. At the same time, the results also indicate that adaptation of the representation family can influence which aspects of downstream performance are favored. In the present experiments, DINOv3 with a frozen decoder achieved slightly higher segmentation overlap, whereas the frozen \ac{ac:BYOS} decoder achieved the lowest segmentation-derived \ac{ac:EF} error. This difference suggests that overlap-based anatomical recovery and downstream functional utility are related but not identical properties of a representation-extraction pipeline.

A further important observation is that segmentation overlap and functional accuracy did not rank methods identically. Across methods, better Dice and IoU generally corresponded to lower \ac{ac:EF} error, but the relationship was not strictly monotonic. The supervised U-Net achieved the highest segmentation overlap, whereas the frozen \ac{ac:BYOS} decoder achieved the lowest segmentation-derived \ac{ac:EF} MAE. Similarly, the strongest frozen and partially fine-tuned DINOv3 variants were separated only modestly in overlap metrics and were not clearly distinguishable in \ac{ac:EF} MAE under paired bootstrap analysis. These findings are consistent with the fact that framewise overlap does not fully capture the functional usefulness of predicted masks. Small contour biases, leakage into adjacent chambers, or modest temporal inconsistency may have limited effect on Dice while still altering the surrogate volume curve used for \ac{ac:EF} estimation. For clinically meaningful downstream assessment, segmentation quality therefore cannot be characterized adequately by overlap metrics alone.

The comparison with direct \ac{ac:EF} prediction further clarifies this point. Direct regression achieved the best learned functional performance overall, confirming that explicit segmentation is not required when the sole objective is accurate \ac{ac:EF} prediction. However, direct regression does not provide anatomically inspectable outputs and therefore offers less support for interpretability, quality control, and downstream measurement. In many medical imaging settings, these properties remain important even when an end-to-end predictor yields lower error on a scalar endpoint. The present results should therefore not be read as an argument against direct functional prediction, but rather as a demonstration that strong segmentation-based pipelines can approach competitive functional performance while preserving anatomical interpretability, and that their quality is best assessed using both anatomical and functional criteria.

The limited gains from partial \ac{ac:FT} are also noteworthy. For both representation families, once a suitable frozen decoder was introduced, unfreezing a restricted subset of backbone or adapter parameters produced only incremental additional benefit. For DINOv3, partial \ac{ac:FT} improved overlap slightly relative to the frozen decoder baseline, but yielded little change in \ac{ac:EF} MAE. For \ac{ac:BYOS}, partial \ac{ac:FT} likewise produced only modest changes relative to the strongest frozen-decoder configuration. This pattern suggests that, in the present setting, the dominant improvement came from increasing the expressivity of the downstream extraction strategy rather than from substantial supervised adaptation of the representation itself. From a practical perspective, this is encouraging: strong frozen-feature transfer may capture much of the attainable benefit while avoiding the additional complexity and computational cost of broader \ac{ac:FT}.

Several limitations should be considered when interpreting these findings. First, all experiments were conducted on a single public benchmark, EchoNet-Dynamic, and were restricted to the apical four-chamber view. Although this dataset provides a useful and widely adopted testbed, the extent to which the observed extraction-strategy dependence generalizes across institutions, acquisition settings, patient populations, or echocardiographic views remains to be established. Second, no external validation dataset was included, so robustness beyond this benchmark has not yet been demonstrated. Third, while the study compared several downstream extraction regimes, it does not exhaust the broader design space of possible downstream adaptations. In particular, temporally explicit decoders, alternative parameter-efficient tuning strategies, or stronger video-native architectures may change the balance between overlap and functional performance. Fourth, \ac{ac:EF} was derived using a simplified area-to-volume surrogate rather than a calibrated volumetric model. The ground-truth-mask ceiling suggests that this proxy was not the dominant source of error in the present experiments, but it remains a simplifying assumption. Finally, the heuristic baselines were designed as restricted no-mask-training extraction regimes intended to probe representational accessibility rather than to maximize label-free segmentation performance, and their absolute performance should be interpreted in that context rather than as a best-case estimate of what label-free segmentation could achieve.

These limitations also indicate several directions for future work. External validation across datasets and views would be important for establishing the robustness of the methodological conclusions. It would also be valuable to extend the evaluation hierarchy to temporally explicit decoders and additional parameter-efficient adaptation strategies. More broadly, the present results suggest that dense \ac{ac:SSL} studies may benefit from treating the evaluation protocol itself as an object of analysis. For dense clinical prediction, conclusions drawn from a single downstream probe may be incomplete, and a more informative picture may emerge only when representations are examined across multiple controlled extraction regimes.

In summary, this study shows that downstream extraction strategy is a major determinant of apparent self-supervised representation quality in dense echocardiographic analysis. Across two complementary representation families, weak heuristic extraction substantially understated what was recoverable from frozen features, whereas lightweight decoder probes revealed much stronger anatomical and functional utility. Limited partial \ac{ac:FT} provided only modest additional gains relative to strong frozen-decoder baselines. These findings support extraction-strategy-aware evaluation as an important methodological consideration for \ac{ac:SSL} in echocardiography and, more cautiously, suggest that similar considerations may apply to other dense medical image analysis settings.

\section{Conclusion}
\label{sec:conclusion}

We presented a systematic evaluation of self-supervised representations for left-ventricular segmentation and \ac{ac:EF} estimation in apical four-chamber echocardiography on EchoNet-Dynamic. The central finding is methodological: in this dense clinical setting, conclusions about representation quality depend strongly on the downstream extraction strategy used for evaluation. Assessing a representation under a single heuristic or probe can therefore provide an incomplete and potentially misleading picture of its practical utility.

Across both representation families studied here, heuristic extraction substantially understated what was recoverable from frozen self-supervised features. More expressive downstream strategies, particularly lightweight decoder probes, revealed markedly stronger anatomical and functional performance, while limited partial \ac{ac:FT} produced only modest additional gains relative to strong frozen-decoder baselines. These results indicate that much of the useful \ac{ac:LV} structure present in strong frozen representations may remain hidden when evaluation is restricted to weak downstream access mechanisms.

Taken together, the findings support multi-strategy, extraction-aware evaluation as an important methodological consideration for \ac{ac:SSL} in dense echocardiographic analysis. More broadly, they suggest that for structured medical image analysis tasks, representation quality is most reliably interpreted in conjunction with the downstream mechanism used to reveal it.

\section{Acknowledgements}
S.M. is a fellow of the AstraZeneca Postdoctoral Research Programme.

\section{Declaration of generative AI and AI-assisted technologies in the writing process}
During the preparation of this work the authors used ChatGPT in order to improve the readability and language of the manuscript. After using this tool, the authors reviewed and edited the content as needed and take full responsibility for the content of the published article.

\bibliographystyle{elsarticle-num-names} 

\bibliography{cas-refs}

\newpage
\appendix
\section{Reproducibility details}
\paragraph{Data and preprocessing}

All experiments used the EchoNet-Dynamic dataset with official split assignments taken directly from \texttt{FileList.csv}. Raw videos were read from the \texttt{Videos} directory, clinical metadata including ejection fraction were read from \texttt{FileList.csv}, and traced endocardial contours were read from \texttt{VolumeTracings.csv}. Unless otherwise stated, training, validation, and test partitions followed the split labels in \texttt{FileList.csv} without additional resplitting.

For mask-supervised segmentation experiments, binary LV cavity masks were generated from the tracing coordinates in \texttt{VolumeTracings.csv}. For each traced frame, the contour point pairs were converted into a closed polygon and rasterized onto a $112 \times 112$ binary mask. When a larger target resolution was required, this binary mask was resized to the model input resolution using nearest-neighbor interpolation. Video frames were read with OpenCV, converted from BGR to RGB, resized to the target spatial resolution using area interpolation, and converted to floating-point intensity values in the range $[0,1]$ by division by 255. No dataset-level mean/std normalization was applied.

Preprocessing differed slightly between BYOS pretraining and downstream supervised training. During BYOS pretraining, all videos in the \texttt{Videos} directory were indexed frame-by-frame with frame stride 1, so every available frame contributed a candidate unlabeled training sample. Each selected frame was resized and center-cropped to $512 \times 512$ before augmentation. In contrast, supervised segmentation baselines and frozen/partial-fine-tuning probe experiments trained on traced labeled frames only, with each training item corresponding to a single traced frame and its derived binary mask. During evaluation of segmentation-derived methods, all frames in each video were processed, typically in temporal chunks of 32 frames for memory efficiency.

For the direct EF regressor baseline, training used clips of 32 frames at $112 \times 112$ resolution sampled as random contiguous temporal windows with stride 2. Evaluation used 32 uniformly sampled frames across the full video. Method-specific dataset usage and preprocessing settings are summarized in Tables~S2A and~S2B.

When cone masking was enabled, the ultrasound field-of-view mask was derived automatically from the video frames themselves. Selected frames were converted to grayscale using $0.299R + 0.587G + 0.114B$, and a pixelwise temporal median image was computed across those frames. The median image was thresholded at a chosen percentile (default 40th percentile) to obtain a foreground support mask. This mask was then morphologically closed with a disk of radius 9, hole-filled, and cleaned by removal of small connected components with minimum size $\max(50,\ 0.001HW)$ for image size $H \times W$. The largest connected component was retained as the cone candidate. If its area was smaller than 20\% of the image area, the cone mask defaulted to the full image. When configured, the cone mask was further eroded by 10 pixels to obtain an inner cone mask used for final prediction restriction.

\begin{table*}[!h]
\centering
\scriptsize
\setlength{\tabcolsep}{3pt}
\renewcommand{\arraystretch}{1.05}
\caption{Compact dataset usage and frame-sampling policy by method family.}
\label{tab:s2a_dataset_usage}
\begin{tabularx}{\textwidth}{
>{\raggedright\arraybackslash}p{2.8cm}
>{\raggedright\arraybackslash}p{1.8cm}
>{\raggedright\arraybackslash}p{1.6cm}
>{\raggedright\arraybackslash}X
>{\raggedright\arraybackslash}p{2.4cm}}
\toprule
\textbf{Method family} & \textbf{Supervision} & \textbf{Train unit} & \textbf{Training sampling} & \textbf{Evaluation usage} \\
\midrule
BYOS pretraining &
none &
single frame &
All videos indexed frame-by-frame with stride 1; every available frame eligible &
Not applicable to pretraining \\

Heuristic baselines &
none &
none &
No supervised training &
All video frames processed at inference \\

Frozen probes &
traced LV masks &
single traced frame &
All traced frames in the selected split; one labeled frame per training item &
All video frames processed at inference in 32-frame chunks \\

Partial fine-tuning &
traced LV masks &
single traced frame &
All traced frames in the selected split; one labeled frame per training item &
All video frames processed at inference in 32-frame chunks \\

Supervised U-Net &
traced LV masks &
single traced frame &
All traced frames in the selected split; one labeled frame per training item &
All video frames processed at inference in 32-frame chunks \\

Direct EF regressor &
clinical EF &
video clip &
Random contiguous 32-frame clips with stride 2 &
32 uniformly sampled frames across the full video \\
\bottomrule
\end{tabularx}
\end{table*}

\begin{table*}[!h]
\centering
\scriptsize
\setlength{\tabcolsep}{3pt}
\renewcommand{\arraystretch}{1.05}
\caption{Compact preprocessing, resizing, and label-generation details.}
\label{tab:s2b_preproc}
\begin{tabularx}{\textwidth}{
>{\raggedright\arraybackslash}p{2.8cm}
>{\raggedright\arraybackslash}p{1.8cm}
>{\raggedright\arraybackslash}p{2.0cm}
>{\raggedright\arraybackslash}p{2.0cm}
>{\raggedright\arraybackslash}X}
\toprule
\textbf{Component} & \textbf{Source} & \textbf{Resize / crop} & \textbf{Interpolation / scaling} & \textbf{Details} \\
\midrule
Frames for BYOS pretraining &
RGB video frame &
Resize + center crop to $512^2$ &
\texttt{ToTensor()} to $[0,1]$ &
Ultrasound-specific two-view augmentation applied after resizing/cropping \\

Frames for segmentation models &
RGB video frame &
Direct resize, typically to $512^2$ &
OpenCV area interpolation; divide by 255 &
Frames read with OpenCV, converted BGR$\rightarrow$RGB, no mean/std normalization \\

Frames for direct EF regressor &
RGB video frame &
Direct resize to $112^2$ &
OpenCV area interpolation; divide by 255 &
Framewise input to CNN-based temporal EF model \\

Tracing-derived masks &
Polygon from \texttt{VolumeTracings.csv} &
Rasterize at $112^2$, then resize if needed &
Nearest-neighbor interpolation &
Closed polygon from traced coordinate pairs; binary LV cavity mask used as supervision \\

Cone mask &
Temporal median grayscale support &
Computed at current image size &
Percentile threshold + morphology &
40th-percentile threshold, closing radius 9, hole filling, small-object removal, largest component, optional erosion by 10 px \\
\bottomrule
\end{tabularx}
\end{table*}

\paragraph{Implementation details.}
For readability, implementation settings are grouped by method family in Tables~\ref{tab:s1a_ssl_heur}--\ref{tab:s1d_supervised_directef}. Unless otherwise stated, segmentation-derived methods used threshold 0.5, minimum object size 300 px, closing radius 3, opening radius 0, inference in 32-frame chunks, and cone masking when enabled. Mask-supervised segmentation models used random horizontal flip ($p=0.5$) and brightness/contrast jitter ($\pm 0.10$).

\begin{table*}[!h]
\centering
\scriptsize
\setlength{\tabcolsep}{3pt}
\renewcommand{\arraystretch}{1.05}
\caption{Self-supervised and heuristic methods.}
\label{tab:s1a_ssl_heur}
\begin{tabularx}{\textwidth}{
>{\raggedright\arraybackslash}p{2.6cm}
>{\centering\arraybackslash}p{0.9cm}
>{\raggedright\arraybackslash}p{2.0cm}
>{\raggedright\arraybackslash}X
>{\raggedright\arraybackslash}p{2.2cm}}
\toprule
\textbf{Method} & \textbf{Input} & \textbf{Frozen / trainable} & \textbf{Core setup} & \textbf{Objective / selection} \\
\midrule
BYOS pretraining &
$512^2$ &
Frozen: DINOv3 S/16. Trainable: adapter, U-Net dec., proj., pred. &
Tok. grid $32\times32\times384$; adapter $1\times1$ + 3 s2 convs; dec. ch. $(256,128,64,32)$; dense out 5; proj./pred. hid. 2048, out 256; AdamW, LR $10^{-3}$, WD $10^{-3}$, bs 2, cosine schedule, clip 1.0, mixed precision, seed 42, mom. 0.95. &
BYOL cosine. Selection: epiplexity + loss trend + preview inspection. \\

DINOv3 heuristic &
$512^2$ &
Frozen: DINOv3 S/16. Trainable: none. &
Patch darkness with optional center and temporal-var. priors; bilinear upsampling; thresholding, cone masking, morphology. &
No supervised training. Validation tuning. \\

BYOS heuristic &
$512^2$ &
Frozen: BYOS ckpt. Trainable: none. &
Frozen dense output with heuristic cavity scoring, thresholding, morphology, cone masking, temporal tracking. &
No supervised training. Validation tuning. \\
\bottomrule
\end{tabularx}
\end{table*}

\begin{table*}[!h]
\centering
\scriptsize
\setlength{\tabcolsep}{3pt}
\renewcommand{\arraystretch}{1.05}
\caption{Frozen probe baselines. Shared settings: AdamW, 20 epochs, batch size 16, BCE + soft Dice, validation Dice.}
\label{tab:s1b_frozen_probes}
\begin{tabularx}{\textwidth}{
>{\raggedright\arraybackslash}p{3.0cm}
>{\centering\arraybackslash}p{0.9cm}
>{\raggedright\arraybackslash}p{1.8cm}
>{\raggedright\arraybackslash}p{1.5cm}
>{\raggedright\arraybackslash}X
>{\centering\arraybackslash}p{0.9cm}
>{\centering\arraybackslash}p{0.9cm}}
\toprule
\textbf{Method} & \textbf{Input} & \textbf{Frozen} & \textbf{Trainable} & \textbf{Core setup} & \textbf{LR} & \textbf{WD} \\
\midrule
BYOS linear probe &
$512^2$ &
BYOS backbone &
$1\times1$ head &
Hook: \texttt{cnn.decoder.final\_conv}; linear $1\times1$ head; bilinear upsampling if needed. &
$10^{-3}$ &
$10^{-4}$ \\

BYOS dec. probe (adapt) &
$512^2$ &
BYOS backbone &
small dec. head &
Hook: \texttt{cnn.encoder.adapt}; hid. 128; 4 upsampling stages; conv-BN-ReLU; drop. 0.10; final $1\times1$. &
$10^{-3}$ &
$10^{-4}$ \\

BYOS dec. probe (dec3) &
$512^2$ &
BYOS backbone &
small dec. head &
Hook: \texttt{cnn.decoder.blocks.3}; hid. 128; 3 upsampling stages; conv-BN-ReLU; drop. 0.10; final $1\times1$. &
$10^{-3}$ &
$10^{-4}$ \\

DINOv3 linear probe &
$512^2$ &
DINOv3 S/16 &
$1\times1$ head &
Frozen patch-token grid; linear $1\times1$ head; bilinear upsampling. &
$10^{-3}$ &
$10^{-4}$ \\

DINOv3 dec. probe &
$512^2$ &
DINOv3 S/16 &
small dec. head &
Frozen patch-token grid; hid. 128; 4 upsampling stages; conv-BN-ReLU; drop. 0.10; final $1\times1$. &
$10^{-3}$ &
$10^{-4}$ \\
\bottomrule
\end{tabularx}
\end{table*}

\begin{table*}[!h]
\centering
\scriptsize
\setlength{\tabcolsep}{4pt}
\renewcommand{\arraystretch}{1.05}
\caption{Partial fine-tuning baselines. Shared settings: $512^2$ input, AdamW, 20 epochs, batch size 16, BCE + soft Dice, validation Dice. All used the same small decoder-head family as the frozen decoder probes (hidden width 128, bilinear upsampling, dropout 0.10, final $1\times1$ output layer).}
\label{tab:s1c_partial_ft}
\begin{tabularx}{\textwidth}{
>{\raggedright\arraybackslash}p{3.1cm}
>{\raggedright\arraybackslash}X
>{\centering\arraybackslash}p{1.05cm}
>{\centering\arraybackslash}p{1.2cm}
>{\centering\arraybackslash}p{0.85cm}
>{\centering\arraybackslash}p{1.15cm}}
\toprule
\textbf{Method} & \textbf{Unfrozen modules} & \textbf{Head LR} & \textbf{Backb. LR} & \textbf{WD} & \textbf{Selection} \\
\midrule
BYOS partial FT (adapt) &
\texttt{cnn.encoder.adapt} + decoder head &
$10^{-3}$ &
$10^{-4}$ &
$10^{-4}$ &
Val. Dice \\

BYOS partial FT (adapt+dec3) &
\texttt{cnn.encoder.adapt} + \texttt{cnn.decoder.blocks.3} + decoder head &
$10^{-3}$ &
$5\times10^{-5}$ &
$10^{-4}$ &
Val. Dice \\

DINOv3 partial FT (last blk + norm) &
final transformer block + final norm + decoder head &
$10^{-3}$ &
$5\times10^{-5}$ &
$10^{-4}$ &
Val. Dice \\

DINOv3 partial FT (last 2 blk + norm) &
final two transformer blocks + final norm + decoder head &
$10^{-3}$ &
$5\times10^{-5}$ &
$10^{-4}$ &
Val. Dice \\
\bottomrule
\end{tabularx}
\end{table*}

\begin{table*}[!h]
\centering
\scriptsize
\setlength{\tabcolsep}{3pt}
\renewcommand{\arraystretch}{1.05}
\caption{Supervised and direct EF baselines.}
\label{tab:s1d_supervised_directef}
\begin{tabularx}{\textwidth}{
>{\raggedright\arraybackslash}p{2.5cm}
>{\centering\arraybackslash}p{0.9cm}
>{\raggedright\arraybackslash}X
>{\raggedright\arraybackslash}p{1.3cm}
>{\raggedright\arraybackslash}p{1.7cm}
>{\centering\arraybackslash}p{0.8cm}
>{\centering\arraybackslash}p{0.8cm}
>{\raggedright\arraybackslash}p{1.4cm}}
\toprule
\textbf{Method} & \textbf{Input} & \textbf{Core setup} & \textbf{Loss} & \textbf{Opt. / LR / WD} & \textbf{Ep.} & \textbf{Bs} & \textbf{Selection} \\
\midrule
Supervised U-Net &
$512^2$ &
2D U-Net, base 32; enc. widths 32, 64, 128, 256, 512; 4 transposed-conv upsampling stages; final $1\times1$. &
BCE + soft Dice &
AdamW / $10^{-3}$ / $10^{-4}$ &
30 &
16 &
Val. Dice \\

Direct EF regressor &
$112^2$ &
Frame CNN + temporal attention + MLP; feat 256, hid. 256, drop. 0.10; train clips 32 fr. with stride 2; eval on 32 uniformly sampled fr. &
MSE + MAE &
AdamW / $10^{-3}$ / $10^{-4}$ &
30 &
8 &
Val. EF MAE \\
\bottomrule
\end{tabularx}
\end{table*}

\paragraph{Abbreviations.}
dec. = decoder; proj. = projector; pred. = predictor; tok. = token; s2 = stride 2; ch. = channels; hid. = hidden dimension; mom. = momentum decay; var. = variance; backb. = backbone; blk = block; norm = normalization; fr. = frames; bs/Bs = batch size; ckpt. = checkpoint; Val. = validation.

\newpage
\newpage
\section{Supplementary statistical analysis}
\label{sec:supp_bootstrap}

To complement the point estimates reported in the main text, we quantified uncertainty in segmentation and EF performance using nonparametric bootstrap resampling at the video level. One row in the evaluation file corresponded to one video, and resampling was therefore performed over videos rather than frames. This choice reflects the dependence structure of the dataset and avoids inflated confidence from treating multiple frames within the same cine loop as independent observations.

For each method, we computed 5000 bootstrap replicates of the mean Dice, mean IoU, and EF MAE across the EchoNet-Dynamic test videos and report the corresponding 95\% percentile confidence intervals. In addition, for key model comparisons we performed paired bootstrap analysis over the common set of test videos evaluated by both methods. In each replicate, the same resampled set of videos was used for both methods, and the mean paired difference was computed. For overlap metrics, a positive paired difference indicates better performance for the first-listed method; for EF MAE, a negative paired difference indicates lower error for the first-listed method.

The updated bootstrap analysis included partial \ac{ac:FT} of DINOv3 using two restricted unfreezing configurations: the final transformer block plus normalization layer, and the final two transformer blocks plus normalization layer. These additional analyses were included to balance the downstream evaluation hierarchy across representation families and to test whether limited supervised backbone adaptation materially improved over the frozen DINOv3 decoder baseline.

Overall, the bootstrap results reinforced the main pattern of the study. Large improvements were observed when moving from heuristic extraction to learned probes and from linear probes to lightweight decoder probes. For example, relative to the heuristic baseline, the DINOv3 linear probe improved Dice by $0.169$ (95\% CI $0.163$--$0.175$), and moving from the DINOv3 linear probe to the frozen decoder improved Dice by a further $0.053$ (95\% CI $0.050$--$0.055$) together with a large reduction in EF MAE. A similar qualitative pattern was observed for BYOS, where the linear probe substantially improved upon the heuristic baseline and the strongest performance was obtained only under decoder-based or partially fine-tuned extraction strategies.

By contrast, the additional gains from limited partial \ac{ac:FT} were comparatively modest. For DINOv3, unfreezing the final two transformer blocks and normalization layer improved Dice by $0.0046$ (95\% CI $0.0034$--$0.0058$) and IoU by $0.0072$ (95\% CI $0.0056$--$0.0088$) relative to the frozen decoder baseline. However, the corresponding EF MAE difference was small and not clearly distinguishable from zero under paired bootstrap analysis ($0.086$, 95\% CI $-0.166$ to $0.336$). Unfreezing only the final transformer block and normalization layer also improved overlap slightly, but yielded worse EF MAE than the frozen decoder. For BYOS, both partially fine-tuned configurations improved overlap relative to the linear probe, but the frozen BYOS decoder retained lower EF MAE than either partially fine-tuned BYOS setting considered here.

Among the strongest learned segmentation-based methods, differences in functional performance were comparatively small. The DINOv3 frozen decoder and the supervised U-Net were separated in Dice and IoU, but not clearly in EF MAE under paired bootstrap analysis. Likewise, the partially fine-tuned DINOv3 model with the final two blocks unfrozen remained below the supervised U-Net in overlap, while the EF MAE difference between the two methods was not clearly distinguishable from zero. The frozen BYOS decoder achieved lower EF MAE than the frozen DINOv3 decoder, whereas the DINOv3 decoder retained slightly higher segmentation overlap. Taken together, these results support the interpretation that extraction strategy choice is the dominant determinant of apparent representation quality in this setting, whereas limited backbone adaptation contributes only incremental additional benefit.

\begin{table*}[!h]
\centering
\caption{Video-level bootstrap summary of method performance on the EchoNet-Dynamic test set. Values are means with 95\% bootstrap confidence intervals. Both BYOS partial fine-tuning configurations reported in the main text are shown separately.}
\label{tab:bootstrap_summary}
\resizebox{\textwidth}{!}{%
\begin{tabular}{lccc}
\toprule
\textbf{Method} & \textbf{Dice} & \textbf{IoU} & \textbf{EF MAE} \\
\midrule
DINOv3 heuristic & 0.684 [0.677, 0.691] & 0.537 [0.530, 0.544] & 13.01 [12.48, 13.56] \\
DINOv3 + linear probe & 0.853 [0.849, 0.857] & 0.752 [0.747, 0.757] & 14.07 [13.52, 14.65] \\
DINOv3 + decoder & 0.906 [0.903, 0.909] & 0.834 [0.829, 0.838] & 9.65 [9.20, 10.09] \\
Partial FT DINOv3 + decoder (last block + norm) & 0.910 [0.907, 0.913] & 0.840 [0.836, 0.844] & 10.26 [9.82, 10.71] \\
Partial FT DINOv3 + decoder (last 2 blocks + norm) & 0.911 [0.908, 0.914] & 0.841 [0.836, 0.845] & 9.57 [9.11, 10.03] \\
BYOS heuristic & 0.687 [0.676, 0.698] & 0.558 [0.548, 0.568] & 17.83 [17.08, 18.62] \\
BYOS + linear probe & 0.838 [0.834, 0.842] & 0.730 [0.725, 0.736] & 13.49 [12.92, 14.06] \\
BYOS + decoder & 0.902 [0.898, 0.905] & 0.827 [0.823, 0.832] & 8.74 [8.32, 9.16] \\
Partial FT BYOS + decoder (adapt only) & 0.906 [0.903, 0.909] & 0.833 [0.829, 0.838] & 9.71 [9.26, 10.15] \\ 
Partial FT BYOS + decoder (adapt + dec3) & 0.906 [0.902, 0.909] & 0.833 [0.828, 0.837] & 9.27 [8.83, 9.72] \\
Supervised U-Net & 0.915 [0.913, 0.918] & 0.848 [0.844, 0.852] & 9.72 [9.29, 10.16] \\
\bottomrule
\end{tabular}%
}
\end{table*}

\begin{table*}[!h]
\centering
\caption{Paired bootstrap comparisons on the EchoNet-Dynamic test set. Reported values are mean paired differences with 95\% bootstrap confidence intervals. For EF MAE, a negative difference indicates lower error for the first-listed method. Where applicable, BYOS partial fine-tuning configurations are reported separately.}
\label{tab:bootstrap_pairwise}
\resizebox{\textwidth}{!}{%
\begin{tabular}{llccc}
\toprule
\textbf{Method A} & \textbf{Method B} & \textbf{Dice difference} & \textbf{IoU difference} & \textbf{EF MAE difference} \\
\midrule
DINOv3 heuristic & DINOv3 + linear probe & -0.169 [-0.175, -0.163] & -0.216 [-0.222, -0.209] & -1.06 [-1.77, -0.36] \\
DINOv3 + linear probe & DINOv3 + decoder & -0.053 [-0.055, -0.050] & -0.081 [-0.085, -0.078] & 4.42 [3.99, 4.86] \\
DINOv3 + decoder & Partial FT DINOv3 + decoder (last block + norm) & -0.0040 [-0.0047, -0.0032] & -0.0064 [-0.0076, -0.0053] & -0.61 [-0.82, -0.39] \\
DINOv3 + decoder & Partial FT DINOv3 + decoder (last 2 blocks + norm) & -0.0046 [-0.0058, -0.0034] & -0.0072 [-0.0088, -0.0056] & 0.09 [-0.17, 0.34] \\
BYOS heuristic & BYOS + linear probe & -0.151 [-0.161, -0.141] & -0.173 [-0.182, -0.164] & 4.35 [3.54, 5.18] \\
BYOS + linear probe & Partial FT BYOS + decoder (adapt only) & -0.0676 [-0.0704, -0.0648] & -0.1029 [-0.1064, -0.0993] & 3.78 [3.33, 4.24] \\
BYOS + decoder & Partial FT BYOS + decoder (adapt only) & -0.0039 [-0.0051, -0.0030] & -0.0058 [-0.0072, -0.0046] & -0.97 [-1.17, -0.77] \\
BYOS + linear probe & Partial FT BYOS + decoder (adapt + dec3) & -0.0673 [-0.0702, -0.0645] & -0.1025 [-0.1061, -0.0988] & 4.22 [3.77, 4.66] \\
BYOS + decoder & Partial FT BYOS + decoder (adapt + dec3) & -0.0036 [-0.0049, -0.0026] & -0.0055 [-0.0069, -0.0041] & -0.54 [-0.73, -0.33] \\
DINOv3 + decoder & Supervised U-Net & -0.0093 [-0.0115, -0.0074] & -0.0143 [-0.0171, -0.0118] & -0.07 [-0.38, 0.25] \\
Partial FT DINOv3 + decoder (last 2 blocks + norm) & Supervised U-Net & -0.0047 [-0.0067, -0.0029] & -0.0072 [-0.0098, -0.0047] & -0.15 [-0.45, 0.15] \\
DINOv3 + decoder & BYOS + decoder & 0.0041 [0.0031, 0.0053] & 0.0061 [0.0048, 0.0076] & 0.92 [0.71, 1.12] \\
Partial FT DINOv3 + decoder (last 2 blocks + norm) & Partial FT BYOS + decoder (adapt only) & 0.0048 [0.0036, 0.0060] & 0.0075 [0.0058, 0.0091] & -0.14 [-0.38, 0.10] \\
Partial FT DINOv3 + decoder (last 2 blocks + norm) & Partial FT BYOS + decoder (adapt + dec3) & 0.0050 [0.0039, 0.0063] & 0.0078 [0.0062, 0.0095] & 0.30 [0.05, 0.54] \\
\bottomrule
\end{tabular}%
}
\end{table*}

\end{document}